\documentclass[10pt,letterpaper]{article}
\usepackage[margin=1in]{geometry}
\usepackage[T1]{fontenc}
\usepackage{times}
\usepackage[nopatch=eqnum]{microtype}
\usepackage{natbib}
\setcitestyle{authoryear,round,citesep={;},aysep={,},yysep={;}}
\renewcommand{\topfraction}{0.95}
\renewcommand{\bottomfraction}{0.8}
\renewcommand{\textfraction}{0.05}
\renewcommand{\floatpagefraction}{0.8}
\makeatletter
\renewcommand{\@maketitle}{%
  \begin{center}
  {\LARGE\bfseries\@title\par}
  \vspace{0.75em}
  {\normalsize\begin{tabular}[t]{c}\@author\end{tabular}\par}
  \end{center}
  \vspace{0.25em}}
\makeatother

\usepackage{amsmath,amsfonts,bm}

\def\eqref#1{equation~\ref{#1}}

\def\1{\bm{1}}

\DeclareMathAlphabet{\mathsfit}{\encodingdefault}{\sfdefault}{m}{sl}
\SetMathAlphabet{\mathsfit}{bold}{\encodingdefault}{\sfdefault}{bx}{n}

\PassOptionsToPackage{hyphens}{url}
\usepackage[hidelinks]{hyperref}
\usepackage{url}
\usepackage{amsmath}
\usepackage{booktabs}
\usepackage{tabularx}
\usepackage{multirow}
\usepackage{graphicx}
\usepackage{capt-of}
\usepackage{xcolor}
\usepackage[normalem]{ulem}
\newif\ifauthorreview
\authorreviewfalse
\definecolor{authorred}{RGB}{180,25,25}

\graphicspath{{figs/}}

\title{I2O2I-Nav: A Benchmark with a Modular Agent for Long-Horizon Indoor-to-Outdoor-to-Indoor Navigation}

\author{%
\small Zhengjie Shu\textsuperscript{1,2,*}\quad
Wanyu Li\textsuperscript{2,*}\quad
Di Xiao\textsuperscript{2,*}\quad
Lingxin Jiang\textsuperscript{5}\quad
Lai Jiang\textsuperscript{2}\quad
Kaidi Hu\textsuperscript{2}\quad
Xincheng Tang\textsuperscript{2}\\[2pt]
\small Wenkang Hu\textsuperscript{2}\quad
Haoran Wang\textsuperscript{2}\quad
Jinchuan Zhang\textsuperscript{1,4}\quad
Chang Chen\textsuperscript{1}\quad
Wei Li\textsuperscript{3}\quad
Jia Pan\textsuperscript{1,\ensuremath{\dagger}}\quad
Ruigang Yang\textsuperscript{2,4,\ensuremath{\dagger}}\\[5pt]
\small\textsuperscript{1}University of Hong Kong\enspace
\textsuperscript{2}Shanghai Jiao Tong University\enspace
\textsuperscript{3}Nanjing University\enspace
\textsuperscript{4}NeoWa Robotics\enspace
\textsuperscript{5}Independent Researcher\\[3pt]
\footnotesize\textsuperscript{*}Co-first authors (equal contribution).\quad
\textsuperscript{\ensuremath{\dagger}}Corresponding authors.\\[3pt]
\small Project Page: \url{https://i2o2i-nav.github.io/}}

\date{}
\begin{document}

\raggedbottom
\maketitle

\par\smallskip
\noindent\begin{minipage}{\textwidth}
\centering
\includegraphics[width=1.0\linewidth]{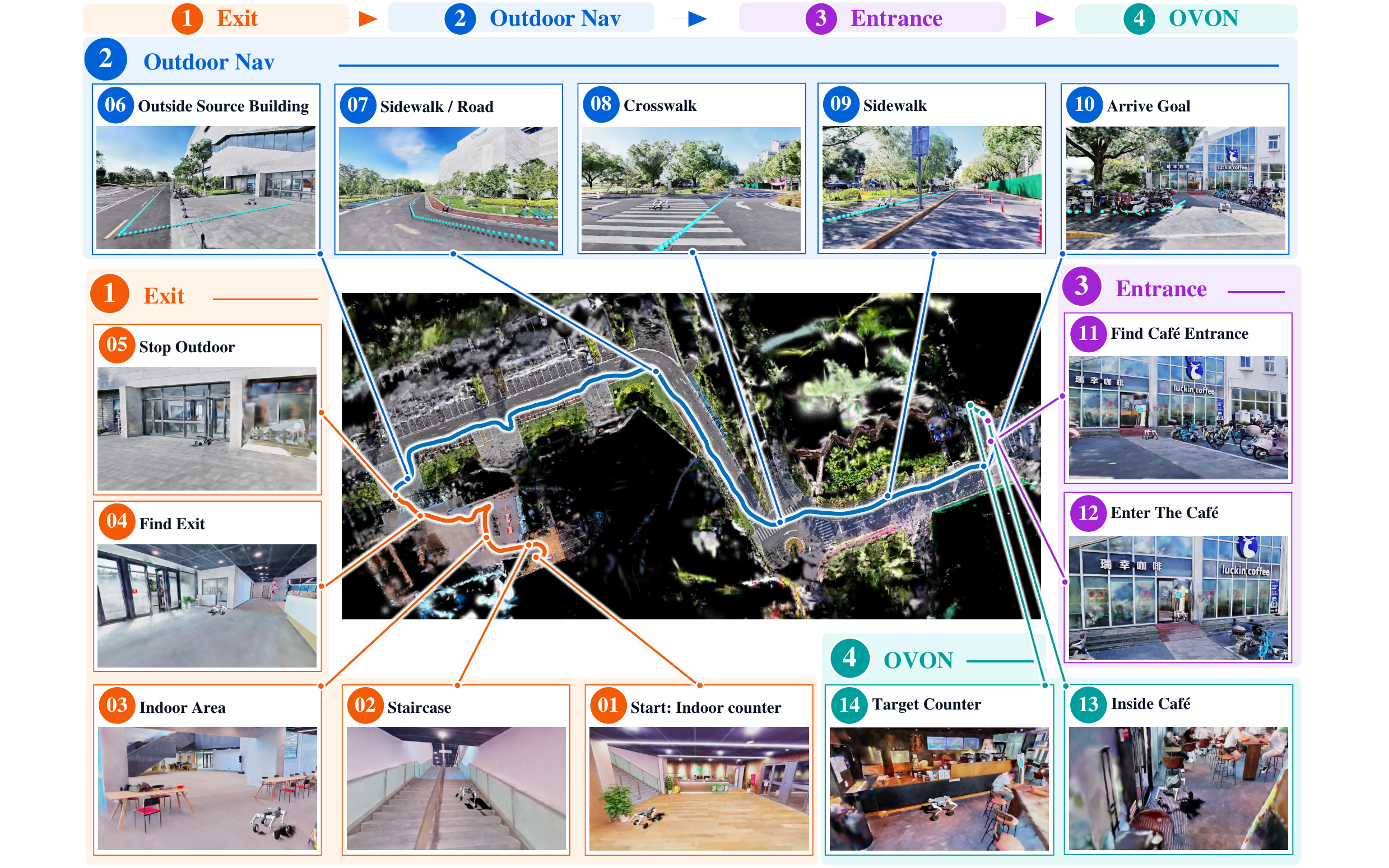}
\captionof{figure}{An I2O2I mission with four stages: Exit, OutdoorNav, Entrance, and OVON.}
\label{fig:cover}
\end{minipage}
\par\smallskip

\begin{abstract}
Service and delivery robots often need to complete long-horizon navigation missions spanning indoor and outdoor environments. Existing embodied navigation benchmarks primarily evaluate isolated stages or partial cross-domain journeys, providing limited support for assessing how heterogeneous navigation capabilities compose under continuous execution. We introduce \textbf{I2O2I-Nav}, a physics-enabled benchmark of 20 real-world composite scenes spanning over $115{,}000\,\mathrm{m}^2$ across multiple buildings and floors, with aligned 3DGS rendering and collision geometry. Its three-level evaluation progresses from independent primitives to two-stage compositions and complete four-stage I2O2I missions. Each stage starts from its predecessor's actual endpoint, while stage-conditional metrics track completion and localize failures along the mission. We further propose the \textbf{MODULAR-Agent} that coordinates frozen navigation policies through completion verification, stage handoffs, and bounded recovery.
Evaluation across 1,374 episodes highlights the difficulty of continuous composition. The MODULAR-Agent achieves 40.5--76\% success on individual primitives, but full I2O2I success remains low. Compared with a fixed policy chain using the same underlying policies, it more than doubles success on both Indoor-to-Outdoor (I2O) and Outdoor-to-Indoor (O2I), while improving full-mission success from 1.15\% to 4.02\%. Stage-wise analysis shows that the MODULAR-Agent enables more missions to reach the final object-navigation stage, where conditional success remains low, revealing challenges in both long-horizon stage execution and completion decisions.

\end{abstract}

\section{Introduction}
\label{sec:introduction}

Service and delivery robots often need to navigate through several distinct environments to complete a single request. For example, a robot asked to ``go to Starbucks and pick up a coffee'' may need to leave its current building, travel along outdoor pedestrian routes, locate an entrance to the destination building, and find the coffee counter inside. A delivery robot carrying a package between buildings faces a similar journey. These applications require continuous indoor-to-outdoor-to-indoor navigation: each stage must begin from the position and orientation reached during the preceding stage. We study the navigation component of such requests; selecting a store, purchasing an item, and manipulating objects are outside the scope of this benchmark.

Each segment presents different challenges and information needs. To leave a building, the robot must explore an unfamiliar interior, potentially traverse floors, locate an exit, and distinguish seeing the outdoors from crossing the boundary. Outdoors, localization and a provided route guide it toward the destination, but it must avoid obstacles and remain within pedestrian-permitted areas. At the destination, the route endpoint may not reveal a usable entrance: the robot must find an opening and enter the correct building. Inside, it must search for and approach the language-specified target, such as a coffee counter. Reliable execution requires individual navigation capabilities, accurate completion decisions, and appropriate stage handoffs. A valid arrival pose for one stage may leave the robot poorly positioned or oriented for the next. Evaluating these capabilities in isolation therefore provides an incomplete picture of reliability across the full journey.

Existing work covers parts of this journey but not its full composition. R2R, RxR, and VLN-CE study indoor language-guided navigation but do not cross to outdoor environments~\citep{anderson2018r2r,ku2020rxr,krantz2020vlnce}. HM3D-OVON addresses open-vocabulary object goals indoors~\citep{hm3dovon}. SocialNav studies socially aware outdoor navigation, while POINav evaluates arrival at precise outdoor points~\citep{socialnav,poinav2026}. BridgeNav crosses the indoor--outdoor boundary in one direction~\citep{bridgenav}. NavVerse evaluates connected indoor-to-outdoor navigation but not the return to a different indoor space~\citep{navverse}. No existing benchmark evaluates how reliably these capabilities compose into a continuous indoor-to-outdoor-to-indoor mission.

We introduce \textbf{I2O2I-Nav} to evaluate this question at three levels of increasing compositional complexity. We make three contributions: 
\textbf{i) I2O2I benchmark.} We introduce a physics-enabled benchmark with 20 large-scale real-world composite scenes for long-horizon indoor-to-outdoor-to-indoor navigation and cover approximately $115{,}602.9\,\mathrm{m}^2$. The scenes combine 3DGS rendering with aligned geometry and include multiple buildings and floors. 
\textbf{ii) Hierarchical evaluation and failure analysis.}
We evaluate four primitives (Exit, OutdoorNav, Entrance, and OVON), two two-stage compositions (I2O and O2I), and full four-stage I2O2I missions. Stage-wise results and qualitative cases identify where and how navigation fails.
\textbf{iii) The MODULAR-Agent.} We propose the MODULAR-Agent (Modular Orchestration with Domain-bridging, Unified Long-horizon, and Adaptive Recovery), a modular and extensible agent that coordinates existing navigation policies for subtask execution, with completion verification, stage handoffs, and bounded recovery, while keeping the underlying policies unchanged.

\section{Related Work}
\label{sec:related}

\subsection{Indoor and Outdoor Embodied Navigation}
R2R and RxR study indoor instruction following, while VLN-CE extends it from navigation graphs to continuous environments~\citep{anderson2018r2r,ku2020rxr,krantz2020vlnce}. VLN-PE examines physical execution, and HM3D-OVON evaluates open-vocabulary object-goal navigation~\citep{vlnpe,hm3dovon}. Outdoors, SocialNav studies socially aware navigation, while POINav evaluates final-meters arrival at precise POI access points~\citep{socialnav,poinav2026}. BridgeNav addresses instruction-guided navigation across the outdoor--indoor boundary~\citep{bridgenav}. These works cover individual capabilities or transitions, rather than a complete mission from an indoor space through outdoor travel to an indoor semantic goal.

\subsection{Cross-Domain Navigation}
Cross-domain navigation studies how navigation systems behave when moving between indoor and outdoor environments. NavVerse evaluates indoor, outdoor, and connected indoor-to-outdoor navigation in continuous robot simulation~\citep{navverse}. Its connected setting covers exit finding, boundary crossing, and subsequent outdoor navigation. This line of work shows that crossing between domains introduces challenges that are not fully captured by evaluating indoor and outdoor navigation separately. However, existing cross-domain benchmarks mainly focus on a single transition or a partial cross-domain journey, rather than the continuous composition of multiple navigation stages across both transition directions.

\subsection{Generalist and Agentic Navigation}
Uni-NaVid, OmniNav, and LightNav-0 develop general-purpose navigation policies, while VLNVerse and OmniNavBench evaluate diverse navigation tasks~\citep{uninavid,omninav,lightnav0,vlnverse,omninavbench}. Agentic approaches add language-model reasoning and map-guided planning (NavGPT, MapGPT), generic instruction following (InstructNav), or skill-based execution (AgentVLN)~\citep{navgpt,mapgpt,instructnav,agentvln}. I2O2I-Nav complements these efforts by evaluating primitive skills and their continuous composition into long-horizon cross-domain missions.
Table~\ref{tab:benchmarks} compares representative benchmarks by environment coverage, task scope, and execution model.

\begin{table}[t]
\centering
\scriptsize
\setlength{\tabcolsep}{2pt}
\renewcommand{\arraystretch}{1.05}
\caption{Comparison of embodied navigation benchmarks. Cross-floor denotes navigation across levels (e.g., via stairs) in benchmark episodes, not merely multi-floor scene assets. $\surd$: included; $\times$: excluded; /: unverified. ``---'' indicates not reported or not applicable.}
\label{tab:benchmarks}
\begin{tabular*}{\textwidth}{@{\extracolsep{\fill}}l c c c l c c@{}}
\toprule
Benchmark & Domain & Cross-floor & 3DGS & Simulator & Robot dynamics & Task dims \\
\midrule
R2R & Indoor & $\surd$ & $\times$ & Matterport3D Simulator & $\times$ & 1 \\
RxR & Indoor & $\surd$ & $\times$ & Matterport3D Simulator & $\times$ & 1 \\
VLN-CE & Indoor & $\surd$ & $\times$ & Habitat & $\times$ & 1 \\
VLN-PE & Indoor & $\times$ & $\surd$ & GRUtopia / Isaac Sim & $\surd$ & 1 \\
HM3D-OVON & Indoor & $\times$ & $\times$ & Habitat & $\times$ & 1 \\
BridgeNav & Outdoor & / & $\times$ & Generated-video evaluation & $\surd$ & 2 \\
POINav & Outdoor & / & $\surd$ & Isaac Sim & $\times$ & 2 \\
VLNVerse & Indoor & / & $\times$ & Isaac Sim & $\surd$ & 5 \\
OmniNavBench & Indoor & $\surd$ & $\times$ & Isaac Sim & $\surd$ & 7 \\
NavVerse & I / O / I2O & / & $\times$ & Isaac Sim & $\surd$ & $\ge$3 \\
\textbf{I2O2I-Nav (ours)} & \begin{tabular}[c]{@{}c@{}}\textbf{I / O / I2O /}\\\textbf{O2I / I2O2I}\end{tabular} & $\boldsymbol{\surd}$ & $\boldsymbol{\surd}$ & \textbf{Isaac Sim / Habitat-GS} & $\boldsymbol{\surd}$ & \textbf{7} \\
\bottomrule
\end{tabular*}
\end{table}

\section{I2O2I-Nav Benchmark}
\label{sec:benchmark}

\subsection{Benchmark Overview}
\label{sec:benchmark-design}

I2O2I-Nav evaluates four navigation primitives: Exit, Outdoor Navigation, Entrance, and OVON. Level~1 tests each independently; Level~2 tests Exit $\to$ Outdoor Navigation (I2O) and Outdoor Navigation $\to$ Entrance (O2I); Level~3 tests the full four-stage sequence. Each primitive may involve multiple behaviors, such as finding stairs and locating an exit.

Composed tasks run continuously without simulator resets: each stage starts from its predecessor's actual position and orientation. A \emph{stage handoff} switches the active task, whereas a \emph{boundary crossing} occurs within Exit or Entrance as the robot moves between indoor and outdoor spaces.

\begin{figure}[t]
\centering
\includegraphics[width=\textwidth]{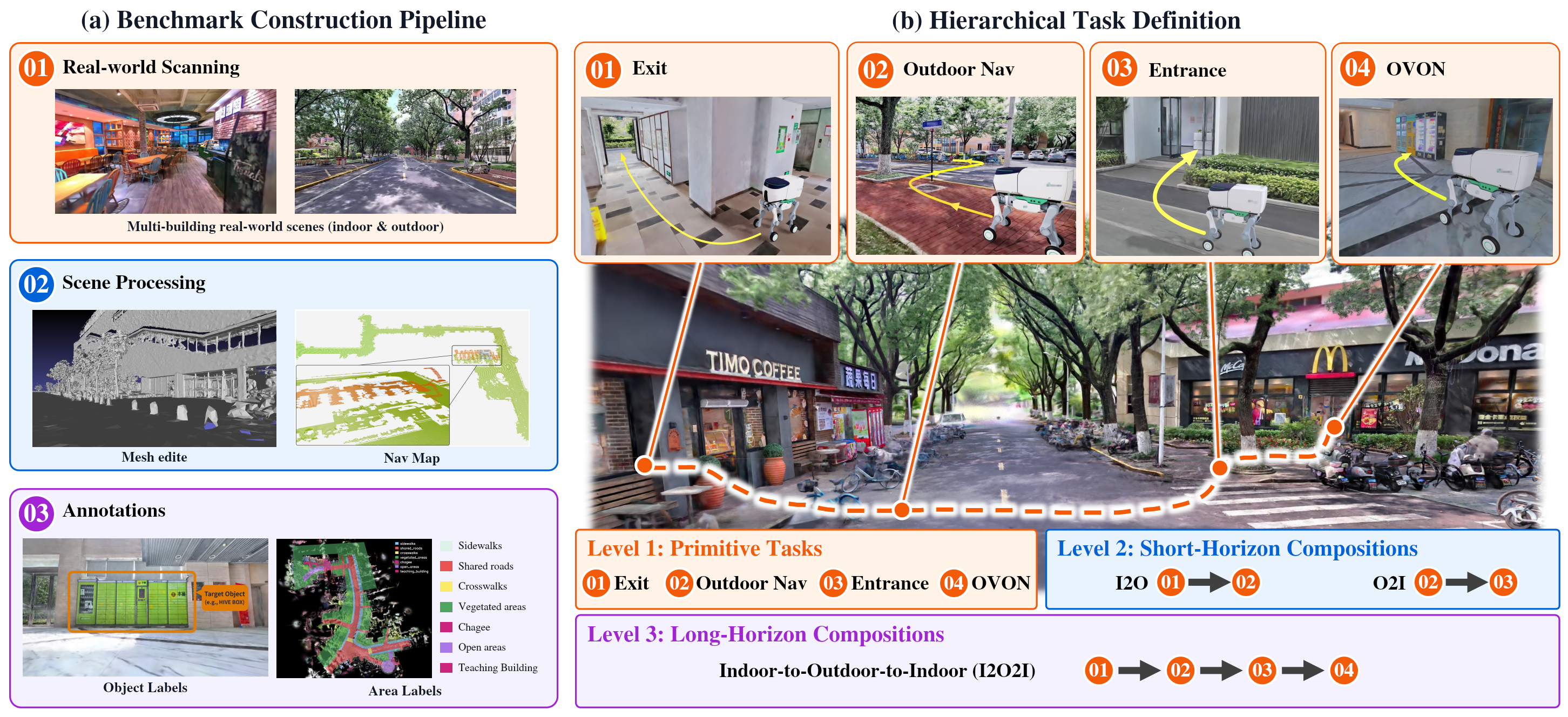}
\caption{
Overview of I2O2I-Nav.
(a) Benchmark construction pipeline. Large-scale real-world indoor--outdoor environments are scanned and processed into executable navigation scenes with collision geometry and navigation maps, together with object-level and area-level semantic annotations.
(b) Hierarchical task definition with increasing compositional complexity. Level~1 evaluates four primitive tasks: Exit, Outdoor Navigation, Entrance, and Open-Vocabulary Object Navigation (OVON). Level~2 evaluates two-stage compositions, Indoor-to-Outdoor (I2O) and Outdoor-to-Indoor (O2I), where the second stage continues from the state reached by the first. Level~3 composes all four primitives into a continuous long-horizon I2O2I mission.
}\label{fig:benchmark-overview}
\end{figure}

\subsection{Scene Construction and Annotation}
\label{sec:scene-construction}

The benchmark contains 20 reconstructed real-world composite scenes with a total annotated area of approximately $115{,}602.9\,\mathrm{m}^2$, covering schools, offices, residences, shopping areas, and parks. The scenes connect indoor and outdoor spaces and include multiple buildings and floors. In Isaac Sim, 3D Gaussians provide visual observations, while aligned meshes support simulation and collision queries. We annotate outdoor region types and pedestrian-access permissions separately to support reference-path planning and region-compliance evaluation. Building, floor, and object labels define task goals and support boundary checks. Evaluator annotations are not exposed to navigation methods. Acquisition, scene processing, and annotation details are provided in Appendix~\ref{app:scene-construction}.

\subsection{Hierarchical Task Definition}
\label{sec:task-definition}
Task predicates and the state-continuity contract are detailed in Appendix~\ref{app:task-definition}.

\par\medskip\noindent\textbf{Level 1: Primitive Tasks}\par\nobreak
Each Level-1 episode evaluates one primitive from a task-specific initial state. Targeting a point goal or area goal. The evaluator checks the corresponding goal and terminal conditions.

\textbf{Exit.} Starting indoors, the robot must find and cross an exit, then issue STOP in a compliant outdoor region. Outdoor regions connect directly and unobstructedly to the open-air pedestrian network, including roofed or covered spaces. Visually open corridors separated from this network by railings, planters, or other barriers remain indoors. Success requires the robot's center to cross the annotated exit boundary and lie within a compliant outdoor region; merely seeing or approaching a doorway is insufficient.
\textbf{Outdoor Navigation.}
The robot starts outdoors and must reach a distant outdoor goal. We evaluate navigation performance through goal-vicinity arrival and trajectory adherence to an optimal reference path, and region-based social compliance through adherence to outdoor regions annotated as compliant with local traversal norms.
\textbf{Entrance.}
The robot starts outdoors near the target building and must find an entrance, cross into that building under the same center-point criterion as Exit, and issue a stop signal at a valid indoor location.
\textbf{Open-Vocabulary Object Navigation (OVON).}
The robot starts indoors with a natural-language object goal and must explore, identify a matching object, navigate to a valid goal location near it, and issue a stop signal. Any matching instance is acceptable.

\par\medskip\noindent\textbf{Level 2: Short-Horizon Composition}\par\nobreak
Level 2 comprises two tasks, each combining two Level-1 primitives. Exit and Entrance require STOP-triggered completion checks, whereas OutdoorNav completes upon reaching the goal tolerance without requiring STOP. Passing the intermediate check advances the task without resetting the robot's pose or environment state. Both stages must be completed for the episode to succeed.

\textbf{Indoor-to-Outdoor (I2O).} The sequence is Exit $\to$ OutdoorNav. The robot starts indoors and must reach a distant outdoor goal. Outdoor localization data and the reference path become available only after successful completion of Exit, modeling scenarios in which outdoor positioning becomes available upon leaving a building. The path starts some distance from the building, requiring the robot to connect its actual exit position to the route.
\textbf{Outdoor-to-Indoor (O2I).} The sequence is OutdoorNav $\to$ Entrance. The robot follows a reference path to the vicinity of the target building, then finds an entrance and enters the building. The path does not specify the entrance, which the robot must locate from its observations.

\par\medskip\noindent\textbf{Level 3: Long-Horizon Composition}\par\nobreak
\textbf{I2O2I} was defined as a long-horizon task that combines four primitives into a continuous episode:
\begin{equation}
\mathrm{Exit} \to \mathrm{OutdoorNav} \to \mathrm{Entrance} \to \mathrm{OVON}.
\end{equation}
The robot starts inside a building and must reach a language-specified object goal inside a different destination building. The task spans two indoor--outdoor transitions and three stage handoffs, following the same stop-triggered completion checks and state continuity rules as Level 2. Success requires completing all four stages, including the final object navigation stage.

\subsection{Episode Generation and Information Conditions}
\label{sec:episode-generation}

We construct Level-1 episodes from indoor, outdoor, and boundary points with semantic annotations. Connectivity checks remove unreachable candidates. Compatible primitives are then linked into Level-2 and Level-3 missions.

Task-provided inputs are distinct from evaluator-only annotations. For compositions beginning with Exit, outdoor localization data and the navigation path become available only after valid Exit completion. Hidden success annotations are withheld from both the policy and its verifier. Appendix~\ref{app:task-generation} details episode records, reference trajectories, stratified sampling, validation, instruction generation, and component-specific information access.

\subsection{Hierarchical Metrics and Failure Localization}
\label{sec:evaluation-protocol}

We evaluate each task level with metrics that capture success, efficiency, progress, and compliance. \textbf{L1 metrics.} Success rate (SR) is the fraction of successful episodes. Success weighted by path length (SPL) measures success-weighted path efficiency. Trajectory length (TL) is mean traveled distance. Distance compliance rate (DCR) is the mean compliant fraction of eligible outdoor travel. Exit, Entrance, and OVON use SR/SPL/TL; OutdoorNav uses SR/SPL/DCR. \textbf{L2/L3 metrics.} Full success rate (FSR) is the fraction completing all stages. Completion success rate (CSR) is the mean fraction of stages completed as a valid ordered prefix. Full-chain SPL weights path efficiency by full success. Route completion (RC) measures normalized progress along the full reference route using the final position's planar projection. DCR retains the outdoor-only definition.

For \textbf{Failure Localization}, stage-conditional success measures stage completion given valid completion of every preceding stage. Together, these metrics capture success, efficiency, progress, compliance, and failure location; Appendix~\ref{app:metrics} provides the formal definitions and evaluation details.

\subsection{The MODULAR-Agent}
\label{sec:agent-method}
\begin{figure}[t]
    \centering
    \includegraphics[width=\linewidth]{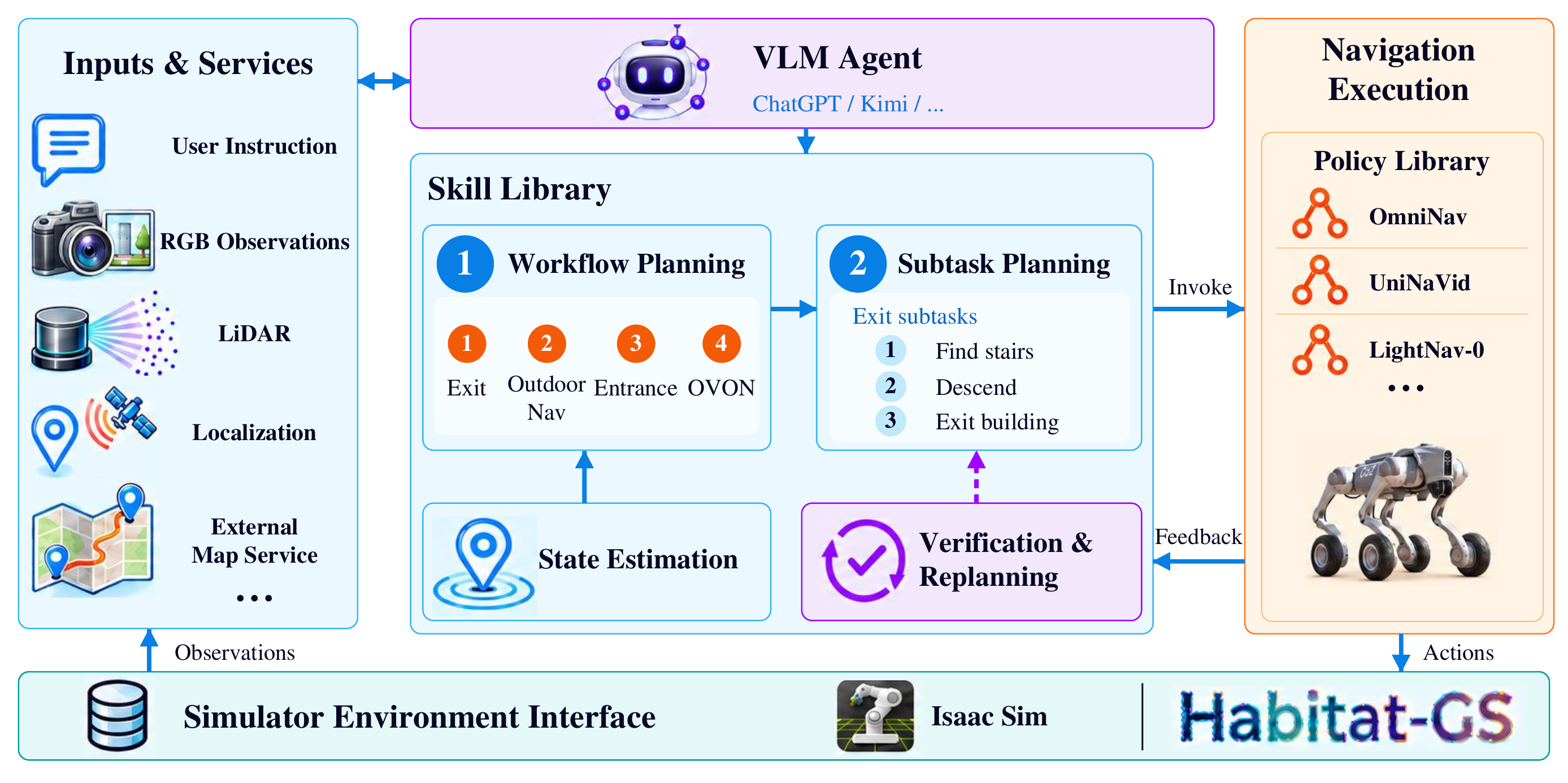}
    \caption{\textbf{Overview of the MODULAR-Agent.} A VLM agent uses instructions, observations, and services to coordinate a skill library and a navigation policy library. State estimation informs workflow planning over Exit, OutdoorNav, Entrance, and OVON. Subtask planning decomposes each workflow into executable steps, illustrated by finding stairs, descending, and exiting a building. Selected navigation policies execute these steps and return feedback for verification and replanning. The simulator environment interface closes the loop by executing actions and returning observations.}
    \label{fig:agent-framework}
\end{figure}

The MODULAR-Agent uses a VLM to orchestrate workflow planning, subtask execution, and verification through reusable skills and navigation policies (Figure~\ref{fig:agent-framework}).

\paragraph{Composable Workflows.}
Exit, OutdoorNav, Entrance, and OVON form reusable workflows. The agent decomposes each into executable subtasks, such as finding stairs, descending, and exiting a building. Benchmark evaluation fixes the workflow order, with execution continuing from the current state without resets.

\paragraph{Navigation Policies Library.}
The library includes OmniNav, Uni-NaVid, LightNav-0, and SocialNav~\citep{omninav,uninavid,lightnav0,socialnav}. Each subtask invokes a compatible frozen policy through an adapter that preserves action semantics. Execution returns feedback for verification and replanning. Evaluated policy assignments appear in Sec.~\ref{sec:experimental-setup}.

\paragraph{Skills Library.}
Four modules close the execution loop: \emph{state estimation} informs \emph{workflow planning}; \emph{subtask planning} generates executable steps; and \emph{verification and replanning} checks execution feedback and revises unsuccessful steps. Verified completion advances execution. Each workflow shares a budget of three replanning cycles across its subtasks, excluding initial planning; failed verification after budget exhaustion ends the workflow. The evaluator independently checks benchmark success. Implementation details appear in Appendix~\ref{app:agent-details}.

\section{Benchmark Evaluation}
\label{sec:experiments}
The evaluation is organized around three research questions. RQ1 concerns independent primitive execution. RQ2 concerns continuation across one stage handoff. RQ3 concerns the full four-stage mission.

\subsection{Experimental Setup}
\label{sec:experimental-setup}
We evaluate in Isaac Sim using a collision-constrained kinematic agent (Appendix~\ref{app:simulation}). Methods share task episodes, observations, action interfaces, success criteria, and execution budgets. Reproducibility details appear in Appendix~\ref{app:interface}; policy/verifier calls, actions, and wall-clock time are reported separately.

\paragraph{Baselines.}
We select task-compatible baselines from Uni-NaVid, OmniNav, SocialNav, and LightNav-0, requiring boundary crossing for Exit/Entrance, reference-route following for OutdoorNav, and language-specified object navigation for OVON. For L1, agent suffixes (O)/(S) denote OmniNav/SocialNav executors, with OmniNav-fast used for OutdoorNav (O) (Table~\ref{tab:l1-primitives}). For L2/L3, (O + S) combines OmniNav for Exit, Entrance, and OVON with SocialNav for OutdoorNav (Tables~\ref{tab:level2} and~\ref{tab:level3}). The fixed chain advances on executor completion proposals; MODULAR-Agent adds completion verification and bounded recovery using the same policies. Benchmark validation checks reference-route connectivity, collision consistency, and reachability through replay, inspects key views with independent completion annotations, and verifies episode targets, floors, and routes.

\subsection{RQ1: Primitive Task Evaluation}
\label{sec:level1}
\textbf{RQ1: How well can current navigation systems solve each primitive independently?}
We evaluate Exit, OutdoorNav, Entrance, and OVON independently using task-compatible methods (Table~\ref{tab:l1-primitives}). The MODULAR-Agent achieves the highest SR in each task, with SRs of 40.50\% on OVON and 76.00\% on OutdoorNav, revealing substantial variation across primitives. Success predicates and metrics follow Sec.~\ref{sec:evaluation-protocol} and Appendices~\ref{app:task-definition} and~\ref{app:metrics}.

\begin{table}[htbp]
\centering
\fontsize{8}{9.6}\selectfont
\setlength{\tabcolsep}{1.5pt}
\renewcommand{\arraystretch}{1.15}
\caption{L1 evaluation. SR, SPL, and DCR are percentages (higher is better); TL is in meters and is descriptive. Agent suffixes (O)/(S) denote OmniNav/SocialNav executors, with OmniNav-fast used for OutdoorNav (O). Bold values indicate the best result per task and metric.}
\label{tab:l1-primitives}
\label{tab:crossing}
\label{tab:exit}
\label{tab:entrance}
\label{tab:ovon}
\label{tab:social}
\begin{tabular*}{\textwidth}{@{\extracolsep{\fill}}l rrr rrr rrr @{\hspace{6pt}}l rrr@{}}
\toprule
& \multicolumn{3}{c}{Exit} & \multicolumn{3}{c}{Entrance} & \multicolumn{3}{c}{OVON} & & \multicolumn{3}{c}{OutdoorNav} \\
\cmidrule(lr){2-4}\cmidrule(lr){5-7}\cmidrule(lr){8-10}\cmidrule(lr){12-14}
Method & SR$\uparrow$ & SPL$\uparrow$ & TL & SR$\uparrow$ & SPL$\uparrow$ & TL & SR$\uparrow$ & SPL$\uparrow$ & TL & Method & SR$\uparrow$ & SPL$\uparrow$ & DCR$\uparrow$ \\
\midrule
Uni-NaVid & 0.00 & 0.00 & 1.14 & 0.00 & 0.00 & 1.49 & 7.00 & 7.00 & 0.59 & OmniNav-fast & 70.00 & 66.15 & 54.50 \\
OmniNav & 29.00 & 18.29 & 156.11 & 14.50 & 10.26 & 35.53 & 38.00 & \textbf{35.73} & 21.19 & SocialNav & 74.00 & 70.23 & 79.39 \\
LightNav-0 & 20.50 & 12.56 & 11.22 & 12.50 & 10.91 & 15.43 & 18.50 & 15.60 & 15.08 & MODULAR-Agent (O) & 74.00 & 69.32 & 55.35 \\
\textbf{MODULAR-Agent (O)} & \textbf{45.50} & \textbf{29.56} & 74.54 & \textbf{51.00} & \textbf{30.78} & 48.02 & \textbf{40.50} & 32.26 & 46.28 & \textbf{MODULAR-Agent (S)} & \textbf{76.00} & \textbf{70.60} & \textbf{80.06} \\
\bottomrule
\end{tabular*}
\end{table}

\subsection{RQ2: Short-Horizon Composition}
\label{sec:level2}
\textbf{RQ2: Can systems execute two adjacent stages without resetting the simulator state?}
I2O composes Exit with OutdoorNav; O2I composes OutdoorNav with Entrance. Each handoff preserves the actual reached state. Table~\ref{tab:level2} reports full-task and stage-conditional performance. With the same underlying policies, the MODULAR-Agent raises I2O FSR from 14.00\% to 30.50\% and O2I FSR from 12.50\% to 32.50\%, more than doubling success in both directions.

\begin{table}[htbp]
\centering
\fontsize{8}{9.6}\selectfont
\setlength{\tabcolsep}{1.5pt}
\renewcommand{\arraystretch}{1.15}
\caption{L2 composition metrics (\%). $C_1=N_1/N$ and $C_2=N_2/N_1$ follow each task's stage order. CSR credits valid prefixes, RC includes partial reference-route progress, and DCR is outdoor-only. Counts and undefined-rate conventions are specified in Appendix~\ref{app:metrics}. Bold values indicate the best result per task and metric.}
\label{tab:level2}
\begin{tabular*}{\textwidth}{@{\extracolsep{\fill}}l rrrrrrr rrrrrrr@{}}
\toprule
& \multicolumn{7}{c}{I2O: Exit $\to$ OutdoorNav} & \multicolumn{7}{c}{O2I: OutdoorNav $\to$ Entrance} \\
\cmidrule(lr){2-8}\cmidrule(lr){9-15}
Configuration & FSR$\uparrow$ & CSR$\uparrow$ & SPL$\uparrow$ & RC$\uparrow$ & DCR$\uparrow$ & $C_1\uparrow$ & $C_2\uparrow$ & FSR$\uparrow$ & CSR$\uparrow$ & SPL$\uparrow$ & RC$\uparrow$ & DCR$\uparrow$ & $C_1\uparrow$ & $C_2\uparrow$ \\
\midrule
OmniNav + SocialNav & 14.00 & 17.25 & 11.90 & 28.50 & 42.24 & 20.50 & 68.29 & 12.50 & 42.75 & 8.78 & \textbf{82.93} & \textbf{53.39} & \textbf{73.00} & 17.12 \\
\textbf{MODULAR-Agent (O + S)} & \textbf{30.50} & \textbf{36.50} & \textbf{23.41} & \textbf{51.32} & \textbf{57.96} & \textbf{42.50} & \textbf{71.76} & \textbf{32.50} & \textbf{52.00} & \textbf{22.27} & 82.43 & 51.92 & 71.50 & \textbf{45.45} \\
\bottomrule
\end{tabular*}
\end{table}

\subsection{RQ3: Long-Horizon I2O2I Composition}
\label{sec:level3}
\textbf{RQ3: How does performance change when all four stages form one continuous mission?}
L3 follows Exit $\to$ OutdoorNav $\to$ Entrance $\to$ OVON without resets and requires completion of the final object goal. Table~\ref{tab:level3} reports full-mission and stage-conditional results. The MODULAR-Agent increases FSR from 1.15\% to 4.02\%, but full-mission completion remains challenging. Section~\ref{sec:diagnostics} analyzes the stage-wise gains and remaining bottlenecks.
\begin{table}[htbp]
\centering
\fontsize{8}{9.6}\selectfont
\setlength{\tabcolsep}{1.5pt}
\renewcommand{\arraystretch}{1.15}
\caption{L3 composition metrics (\%). $C_k=N_k/N_{k-1}$, with $N_0=N$, follows Exit, OutdoorNav, Entrance, and OVON. FSR requires all four stages. RC uses reference-route progress; DCR excludes indoor distance. Bold values indicate the best result per task and metric.}
\label{tab:level3}
\begin{tabular*}{\textwidth}{@{\extracolsep{\fill}}l rrrrr rrrr@{}}
\toprule
& \multicolumn{5}{c}{Full mission} & \multicolumn{4}{c}{Stage-conditional success} \\
\cmidrule(lr){2-6}\cmidrule(lr){7-10}
Configuration & FSR$\uparrow$ & CSR$\uparrow$ & SPL$\uparrow$ & RC$\uparrow$ & DCR$\uparrow$ & $C_1\uparrow$ & $C_2\uparrow$ & $C_3\uparrow$ & $C_4\uparrow$ \\
\midrule
OmniNav + SocialNav & 1.15 & 14.22 & 0.90 & 35.62 & 43.81 & 29.89 & 67.31 & 28.57 & 20.00 \\
\textbf{MODULAR-Agent (O + S)} & \textbf{4.02} & \textbf{26.15} & \textbf{1.08} & \textbf{47.79} & \textbf{56.05} & \textbf{45.40} & \textbf{79.75} & \textbf{52.38} & \textbf{21.21} \\
\bottomrule
\end{tabular*}
\end{table}

\subsection{Failure Analysis and Main Findings}
\label{sec:diagnostics}

\begin{figure}[t]
\centering
\includegraphics[width=\linewidth]{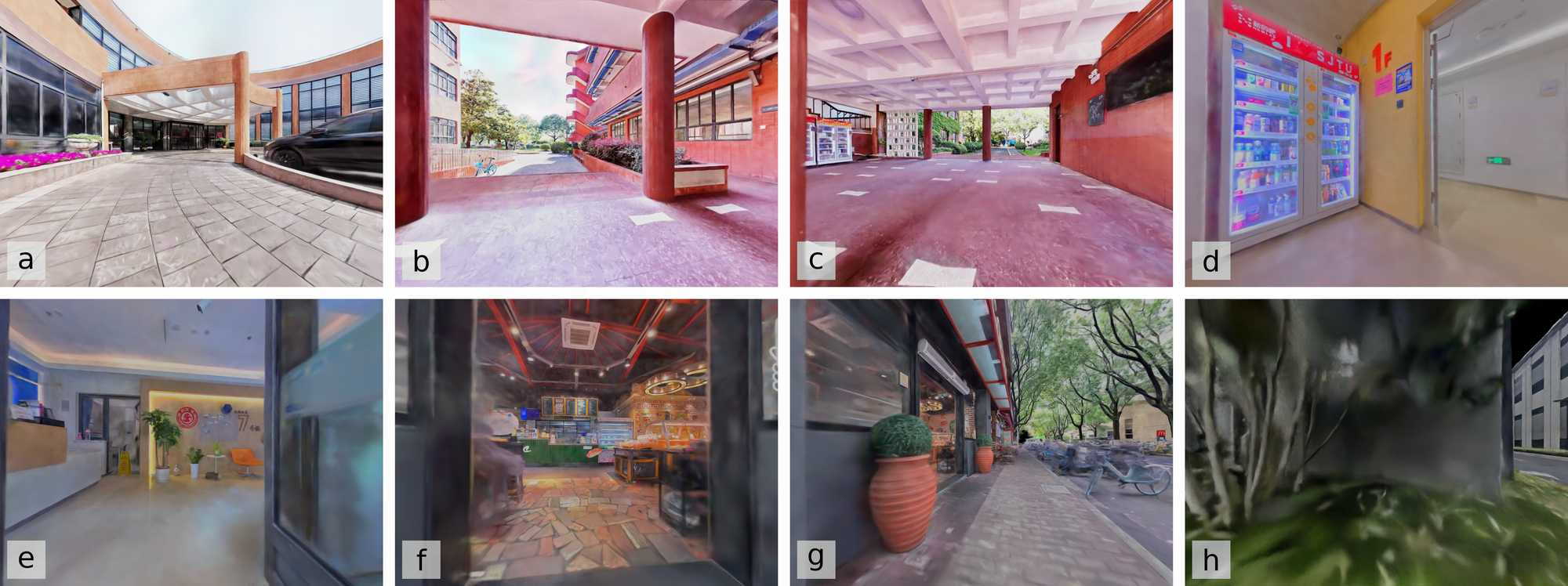}
\caption{Representative L1 failures. Exit: (a) timeout without STOP after reaching outdoors; (b) premature STOP before crossing outside. OVON: (c,d) failure to recognize a visible vending-machine goal. Entrance: (e,f) premature STOP outside a doorway; (g) timeout without entering a visible entrance. OutdoorNav: (h) repeated forward motion into vegetation, resulting in stagnation.}
\label{fig:primitive-failures}
\end{figure}

\paragraph{Primitive failures: recognizing goals and acting on completion.}
Figure~\ref{fig:primitive-failures} highlights recurring failures in goal recognition, completion decisions, and adaptation to stalled progress.

\textbf{Exit: missed and premature stopping.}
In (a), the robot has clearly reached the outdoors but times out without STOP. In (b), it sees outdoor space but stops before leaving the building. These opposite errors expose a mismatch between visual cues and physical task state: successful Exit requires both crossing the boundary and recognizing completion.

\textbf{OVON: visible targets are missed during execution.}
In (c), the requested vending machine appears at the lower left but goes unnoticed. In (d), the robot passes directly beside it without recognizing the goal. Both cases expose a gap between encountering a target and completing navigation: object search must connect recognition to goal-directed approach and timely stopping.

\textbf{Entrance: approaching a doorway is mistaken for entering.}
In (e,f), the robot stops in front of the doorway while still outside. In (g), an indoor entrance is clearly visible ahead to the left, yet the robot fails to enter and times out without STOP. These cases make boundary-aware completion central to cross-domain navigation: approaching an opening must be followed by actual entry and a valid stop.

\textbf{OutdoorNav: persistent actions despite blocked progress.}
In (h), the robot repeatedly advances into vegetation and becomes stuck instead of changing direction. Detecting stagnation and revising execution are therefore essential complements to route following. Together, these cases motivate the MODULAR-Agent's completion verification and bounded replanning around frozen policies.

\paragraph{Short-horizon composition: stronger boundary-task completion.}
The quantitative results identify boundary tasks as a major source of improvement. With the same OmniNav executor, the MODULAR-Agent raises L1 Exit SR from 29.00\% to 45.50\% and Entrance SR from 14.50\% to 51.00\% (Table~\ref{tab:l1-primitives}). These gains of 16.50 and 36.50 percentage points exceed the 2.50-point OVON gain from 38.00\% to 40.50\%, highlighting the agent's stronger benefit on boundary-task completion. This pattern extends to continuous composition (Table~\ref{tab:level2}). In I2O, Exit success increases from 20.50\% to 42.50\%, while conditional OutdoorNav success rises from 68.29\% to 71.76\%; FSR improves from 14.00\% to 30.50\%. In O2I, OutdoorNav success remains comparable at 73.00\% and 71.50\%, but conditional Entrance success increases from 17.12\% to 45.45\%, raising FSR from 12.50\% to 32.50\%. Thus, the agent more than doubles full success in both directions using the same underlying policies. Stage-conditional evaluation localizes these gains to valid exit and entry under the states actually reached during continuous execution.

\paragraph{Long-horizon composition: improved reach, unresolved final goals.}
The MODULAR-Agent increases L3 FSR from 1.15\% to 4.02\%, approximately a 3.5-fold improvement, and CSR from 14.22\% to 26.15\% (Table~\ref{tab:level3}). Conditional success improves across Exit (29.89\% to 45.40\%), OutdoorNav (67.31\% to 79.75\%), and Entrance (28.57\% to 52.38\%), bringing more missions to the final indoor search. Yet OVON conditional success remains at 20.00\% for the fixed chain and 21.21\% for the agent: nearly four fifths of missions reaching this stage fail to complete the object goal. Exit also blocks over half of the agent's missions. Stage-wise evaluation thus distinguishes early attrition from final-goal failure while revealing gains obscured by full success alone.

\paragraph{Success and efficiency: task-dependent benefits.}
Success-weighted path efficiency improves substantially in short compositions: L2 SPL rises from 11.90\% to 23.41\% for I2O and from 8.78\% to 22.27\% for O2I (Table~\ref{tab:level2}). In L1 Exit, higher SR accompanies an SPL increase from 18.29\% to 29.56\% and a reduction in mean TL from 156.11\,m to 74.54\,m. OVON presents a different outcome. The agent achieves the highest SR at 40.50\%, compared with OmniNav's 38.00\%, but SPL decreases from 35.73\% to 32.26\%, a drop of 3.47 percentage points. Mean TL increases from 21.19\,m to 46.28\,m, approximately 2.18 times as long. Thus, its 2.50-point success gain accompanies lower success-weighted efficiency and greater average travel. Since TL includes both successful and failed episodes, it measures overall execution length rather than recovery cost alone. In L3, SPL rises only from 0.90\% to 1.08\% despite the larger FSR gain. Joint evaluation therefore identifies both the benefits of orchestration for boundary tasks and short compositions, and efficient object search as a remaining priority for improving complete missions.

\paragraph{Outdoor compliance: evaluating how the goal is reached.}
I2O2I-Nav further distinguishes successful arrival from adherence to pedestrian-permitted regions. In L1 OutdoorNav, SocialNav and OmniNav-fast achieve SRs of 74.00\% and 70.00\%, but DCRs of 79.39\% and 54.50\%, respectively (Table~\ref{tab:l1-primitives}). Similar arrival rates therefore conceal a large difference in region compliance. With SocialNav, the agent achieves 76.00\% SR and 80.06\% DCR. In composed tasks, the agent improves DCR from 42.24\% to 57.96\% for I2O and from 43.81\% to 56.05\% for L3, alongside higher FSR. O2I shows why both metrics remain necessary: its FSR improves while DCR changes from 53.39\% to 51.92\%. By measuring compliant outdoor travel in both successful and failed episodes, the benchmark captures route behavior that goal-arrival evaluation alone misses.

\section{Conclusion}
\label{sec:conclusion}

\textbf{Limitations}. Manual scene construction limits scale and diversity, while reconstruction artifacts can affect perception and collision checks. Evaluation covers 20 scenes and 1,374 episodes using a collision-constrained kinematic agent. Broader coverage, improved reconstruction, and evaluation with articulated robot dynamics remain future work.

I2O2I-Nav evaluates navigation from standalone primitives to continuous indoor-to-outdoor-to-indoor missions, with stage-wise metrics to localize failures. The MODULAR-Agent coordinates frozen policies through completion verification and bounded recovery, improving full-mission success from 1.15\% to 4.02\%. These results demonstrate progress while identifying reliable stage completion and final object navigation as persistent challenges.

\subsection*{Reproducibility Statement}
The evaluation protocol is described in Sec.~\ref{sec:experiments}. Appendices~\ref{app:scene-construction}--\ref{app:agent-details} document scene construction and annotation, task definitions, episode generation, metrics, simulation settings, and agent execution. Benchmark evaluation code and MODULAR-Agent system code will be made available through our project repository at \url{https://github.com/i2o2i-nav/i2o2i_nav}. We plan to release scene assets, annotations, and associated benchmark data through Hugging Face, with documentation of access and licensing terms. These materials will support reproduction of the benchmark and agent evaluation.

\subsection*{AI Use Statement}
Generative AI tools assisted with research planning and refinement, methodology and experiment design, method implementation and code editing, data annotation and task-instruction preparation, qualitative analysis of navigation traces and images, interpretation of experimental results, literature search and reference checking, figure preparation, and manuscript drafting, translation, and editing. AI-assisted work was reviewed by the authors, who made the final research decisions. Region proposals and object labels were checked and corrected by human annotators (Appendix~\ref{app:region-object-annotation}); instruction preparation is described in Appendix~\ref{app:task-generation}. AI-assisted analysis and writing did not constitute additional experimental runs. Model-based planning and completion verification are components of the evaluated MODULAR-Agent (Appendix~\ref{app:agent-details}), while the evaluator independently determines benchmark success. The authors take responsibility for all final text, code, data, citations, claims, and reported results, including contributions developed with AI assistance.

\subsection*{Ethics Statement}
The necessary permissions have been obtained for scene acquisition, processing, and the intended release of the benchmark assets. The use and redistribution of third-party materials comply with their applicable licenses and permissions. Data release will respect privacy and exclude personal or sensitive information not authorized for disclosure. AI-assisted annotations are reviewed and corrected by human annotators. Reported experiments are conducted in simulation using a collision-constrained kinematic agent; they do not establish safety or locomotion reliability for physical robots. Real-world deployment requires separate validation of navigation, collision avoidance, and robot control under the intended operating conditions.

\clearpage
\bibliography{references}
\bibliographystyle{iclr2027_conference}

\clearpage
\appendix

\begingroup
\raggedbottom
\renewcommand{\topfraction}{0.9}
\renewcommand{\bottomfraction}{0.8}
\renewcommand{\textfraction}{0.08}
\renewcommand{\floatpagefraction}{0.65}
\setcounter{topnumber}{3}
\setcounter{bottomnumber}{2}
\setcounter{totalnumber}{4}
\setlength{\floatsep}{12pt plus 2pt minus 2pt}
\setlength{\textfloatsep}{14pt plus 2pt minus 2pt}
\setlength{\intextsep}{12pt plus 2pt minus 2pt}
\makeatletter
\setlength{\@fptop}{0pt}
\setlength{\@fpsep}{14pt plus 2pt}
\setlength{\@fpbot}{0pt plus 1fil}
\makeatother

\section{Scene Construction}
\label{app:scene-construction}

\subsection{Acquisition and Reconstruction}
We scanned the scenes with an RTK-equipped XGRIDS Lixel K2 handheld 3D scanner. Each scan produced aligned Gaussian and mesh PLY files. The Gaussian representation supplies visual appearance; the mesh supports collision queries. We convert the Gaussian PLY to an Omniverse-compatible USD asset through the 3DGRUT export workflow and import the aligned mesh into the same scene~\cite{nvidia3dgrutexport}. Isaac Sim's RTX renderer supports Gaussian particle fields alongside polygonal content~\cite{isaacsimnurec}.

Figures~\ref{fig:scanned-scenes} and~\ref{fig:scanned-meshes} show Gaussian renderings and mesh views. Table~\ref{tab:scene-inventory} summarizes scene areas, object goals, and environment types.

\begin{figure}[!htbp]
\centering
\includegraphics[trim=2bp 50bp 3bp 50bp,clip,width=\textwidth,height=0.87\textheight,keepaspectratio]{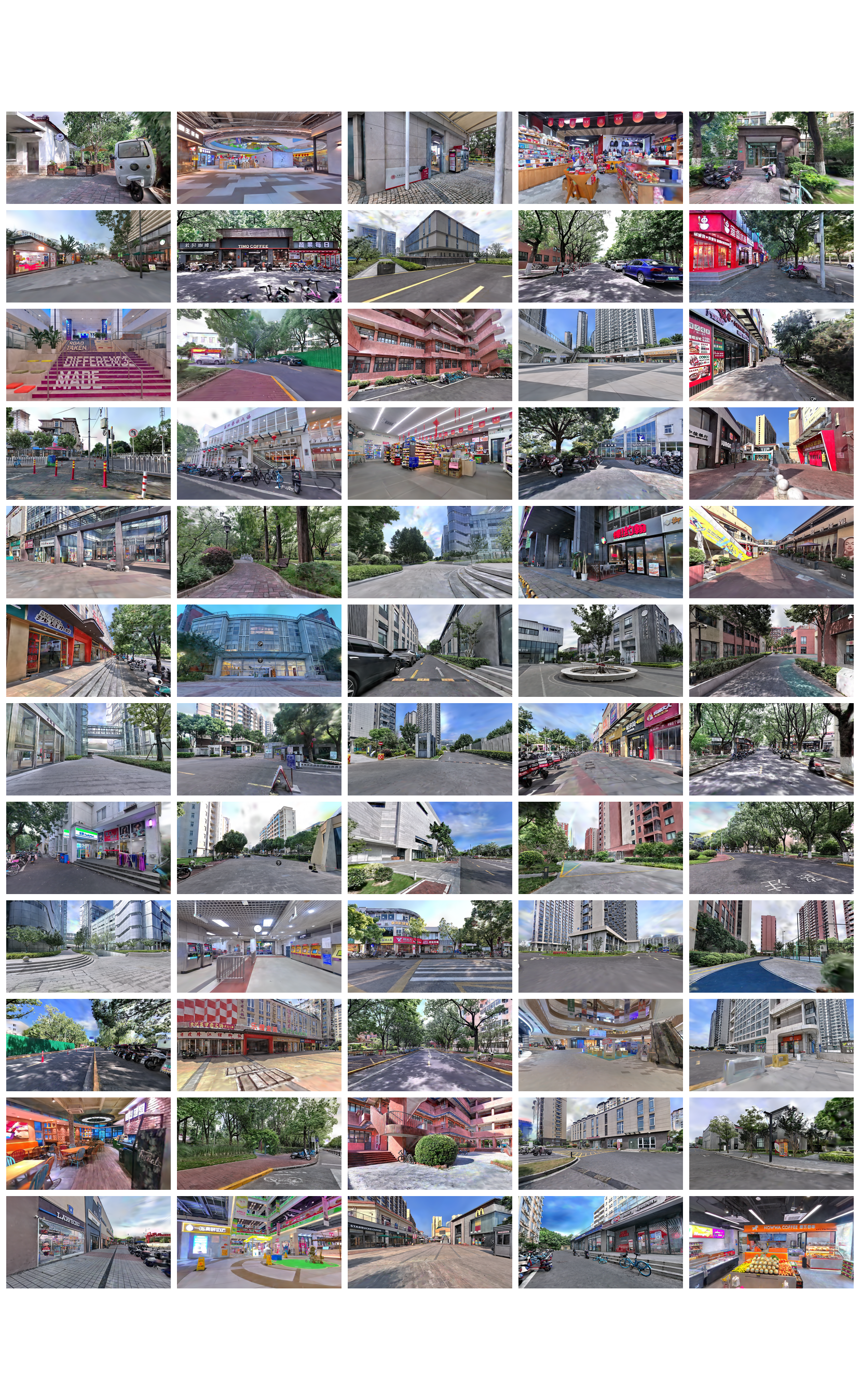}
\caption{3D Gaussian renderings of the scanned scenes.}
\label{fig:scanned-scenes}
\end{figure}

\begin{figure}[!htbp]
\centering
\includegraphics[trim=2bp 38bp 3bp 38bp,clip,width=\textwidth,height=0.30\textheight,keepaspectratio]{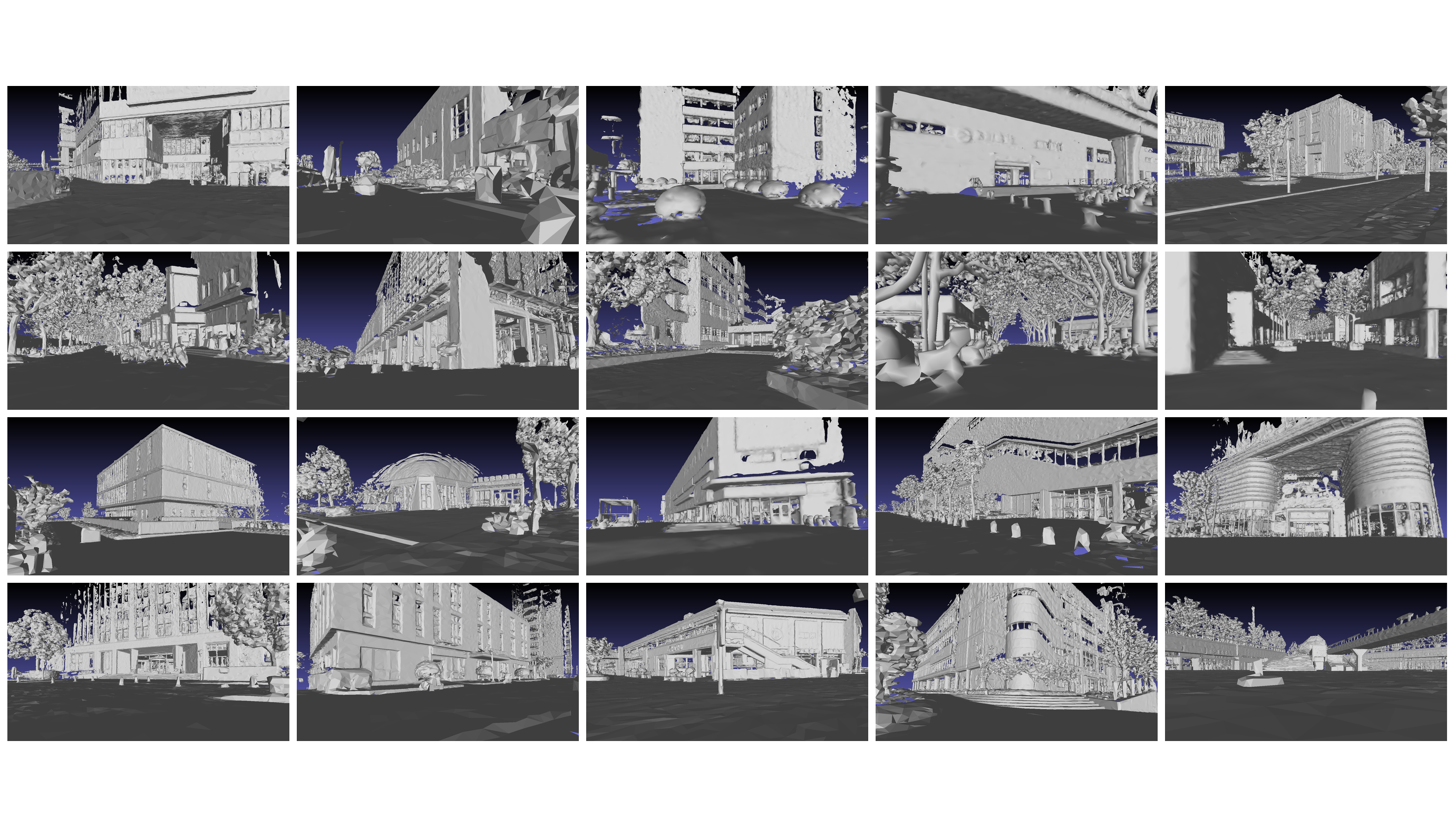}
\caption{Mesh views of the reconstructed geometry used for simulation and collision queries.}
\label{fig:scanned-meshes}
\end{figure}

\begin{table}[!htbp]
\centering\small
\setlength{\tabcolsep}{9pt}
\renewcommand{\arraystretch}{1.03}
\caption{Inventory of the 20 scanned scenes, with a total area of $115{,}602.9\,\mathrm{m}^2$.}
\label{tab:scene-inventory}
\begin{tabular}{@{}cccc@{}}
\toprule
Scene & Area (m$^2$) & Object goals & Environment type \\
\midrule
Scene 1 & 17,027.6 & 8 & School \\
Scene 2 & 5,350.7 & 8 & School \\
Scene 3 & 4,565.1 & 13 & School \\
Scene 4 & 4,508.1 & 10 & School \\
Scene 5 & 4,566.8 & 6 & School \\
Scene 6 & 5,080.6 & 15 & School \\
Scene 7 & 2,625.3 & 6 & School \\
Scene 8 & 5,146.6 & 5 & Office \\
Scene 9 & 2,182.3 & 13 & Office \\
Scene 10 & 5,128.1 & 10 & Residential \\
Scene 11 & 4,498.5 & 24 & Residential \\
Scene 12 & 16,430.5 & 41 & Residential \\
Scene 13 & 4,532.8 & 18 & Residential \\
Scene 14 & 4,585.3 & 13 & Residential \\
Scene 15 & 3,861.3 & 16 & Commercial \\
Scene 16 & 3,139.5 & 10 & Commercial \\
Scene 17 & 8,116.9 & 6 & Commercial \\
Scene 18 & 4,852.4 & 10 & Commercial \\
Scene 19 & 6,470.7 & 13 & Commercial \\
Scene 20 & 2,933.8 & 7 & Park \\
\midrule
\textbf{Total} & \textbf{115,602.9} & \textbf{252} & \textbf{5 types} \\
\bottomrule
\end{tabular}
\end{table}

\subsection{Scene Editing and Processing}
We remove selected Gaussian content in obstructed or poorly reconstructed regions with XGRIDS' editing software~\cite{xgridsmodeleditor}. We remove the corresponding mesh faces in MeshLab~\cite{meshlabsoftware}, keeping visual appearance and collision geometry consistent.

Roadblocks close branches with insufficient scan coverage, particularly at intersections. Each roadblock has both visible geometry and collision geometry. Figure~\ref{fig:scene-roadblocks} shows the four asset types.

\begin{figure}[!htbp]
\centering
\includegraphics[width=0.78\textwidth,height=0.23\textheight,keepaspectratio]{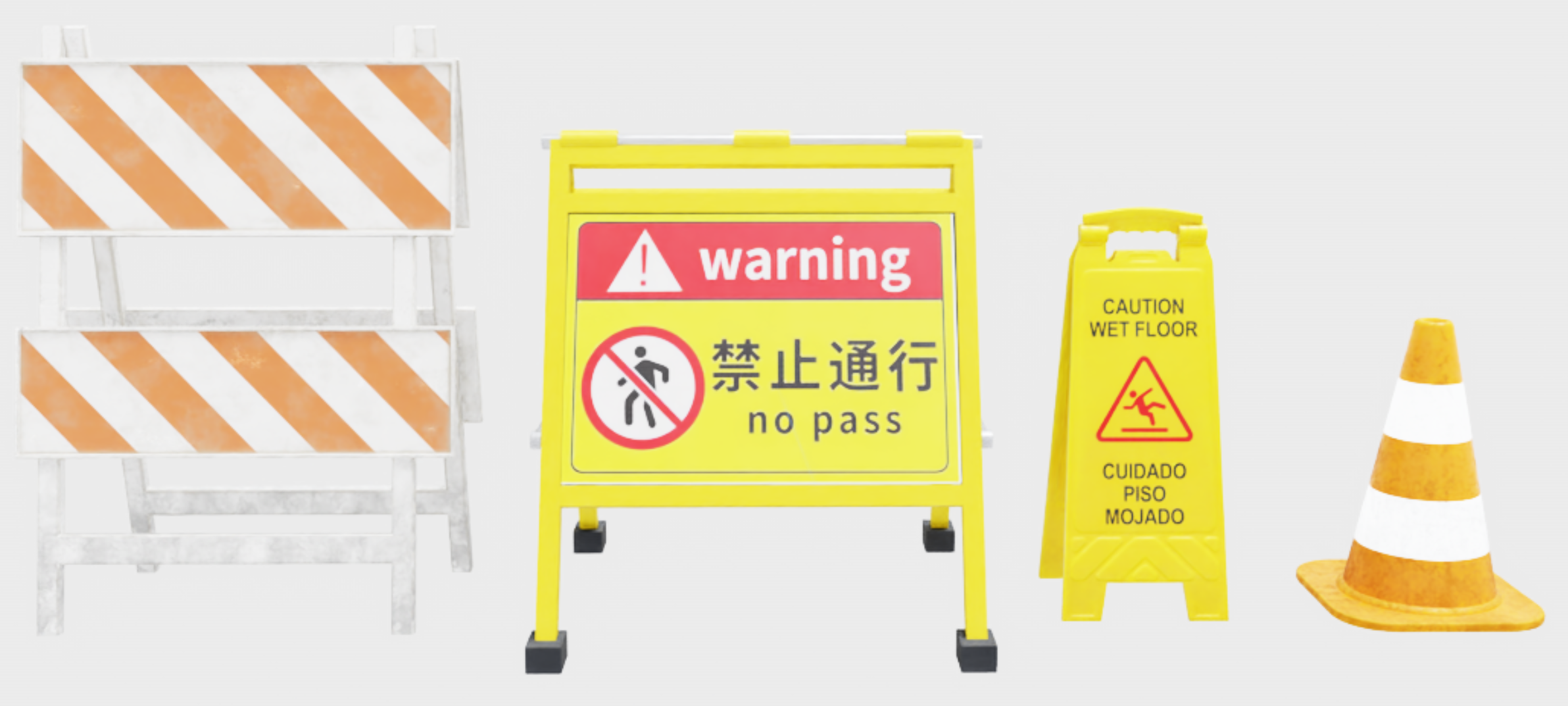}
\caption{Roadblock assets used to close unscanned or poorly reconstructed branches.}
\label{fig:scene-roadblocks}
\end{figure}

After scene editing, we use PCT planner's point-cloud tomography framework~\cite{yang2024efficient} to derive the traversable representation used in task generation. Figure~\ref{fig:scene-nav-mesh} shows the resulting navigation map. Planner settings and reference generation are described in Appendix~\ref{app:task-generation}.

\begin{figure}[!htbp]
\centering
\includegraphics[width=0.88\textwidth,height=0.30\textheight,keepaspectratio]{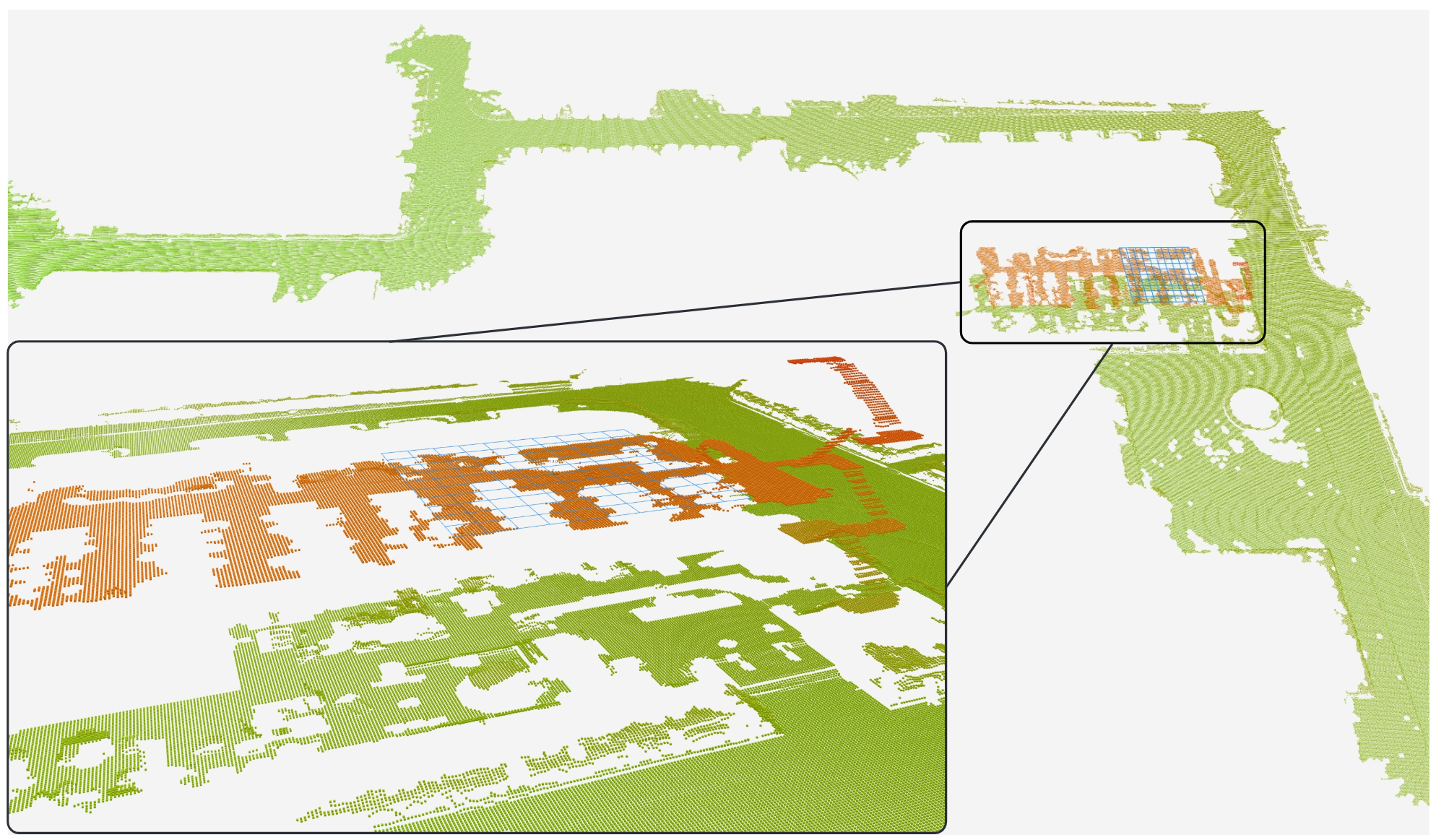}
\caption{Navigation map after scene editing and roadblock placement, with an enlarged local view.}
\label{fig:scene-nav-mesh}
\end{figure}

\subsection{Region and Object Annotation}
\label{app:region-object-annotation}

\paragraph{Region labels.}
Region annotations describe the use of outdoor space and its pedestrian-access status. We distinguish seven region types:
\begin{itemize}
    \setlength{\itemsep}{0pt}
    \item \textbf{Sidewalks:} paths designated primarily or exclusively for pedestrians.
    \item \textbf{Crosswalks:} designated pedestrian crossing areas on roads, including zebra crossings.
    \item \textbf{Shared roads:} roads used by both pedestrians and vehicles, such as internal roads in residential compounds or campuses.
    \item \textbf{Vehicle lanes:} lanes designated primarily for motor vehicles, where walking along the traffic lane is generally not permitted.
    \item \textbf{Open areas:} open spaces that can accommodate pedestrian movement, such as plazas and courtyards.
    \item \textbf{Vegetated areas:} lawns, grass patches, planted strips, and shrub-covered areas.
    \item \textbf{Water areas:} ponds, pools, rivers, and other water surfaces.
\end{itemize}

Region type and pedestrian access are separate attributes. \emph{Allowed} regions include sidewalks, crosswalks, and pedestrian-permitted shared roads, and open areas. \emph{Prohibited} regions include restricted planting areas, vehicle-only lanes, and water areas. Human annotators resolve ambiguous access status and assign final allowed or prohibited labels. Physical traversability and socially permitted movement are distinct, as also emphasized by SocialNav~\cite{socialnav}.

Region proposals generated by Kimi K3 from top-down views are reviewed and corrected by human annotators (Figure~\ref{fig:region-annotation}). Regions are stored as polygons. The labels support reference-path planning and compliance evaluation. Building and floor identities support target-specific boundary checks. Indoor/outdoor boundaries follow the connectivity-based definition in Appendix~\ref{app:task-definition}, rather than roof coverage alone.

\begin{figure}[!htbp]
\centering
\includegraphics[width=0.66\textwidth,height=0.38\textheight,keepaspectratio]{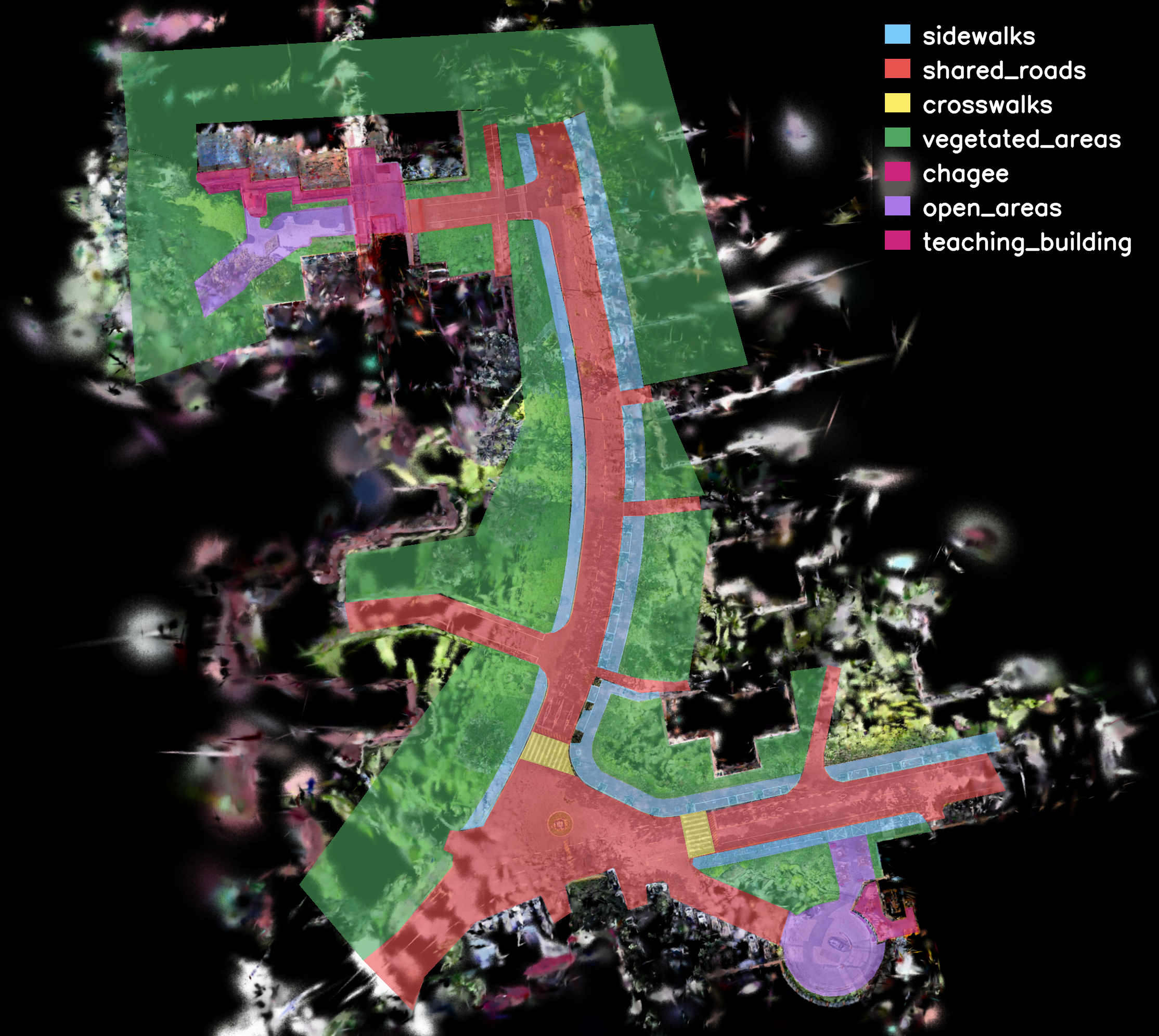}
\caption{Top-down region annotation. Region type and pedestrian-access status are separate attributes.}
\label{fig:region-annotation}
\end{figure}

\paragraph{Object and area goals.}
Object annotations ground language-specified goals in scene instances, supporting open-vocabulary navigation~\cite{hm3dovon}. Building and floor identities distinguish spatially separate targets. When multiple instances satisfy an OVON goal, any accepted instance is valid.

Human annotators review object labels generated by Kimi K3 and add missing targets. Each object instance is stored as a three-dimensional goal point with its semantic label. Figure~\ref{fig:object-goal-annotation} contrasts an object goal with an area goal; valid stopping states follow the task predicates in Appendix~\ref{app:success-predicates}. Instructions use the annotated goal names without synonym variants. Figure~\ref{fig:object-goal-vocabulary} is derived from actual annotated goal names, all of which are admissible OVON targets; it does not report category or episode frequencies.

\begin{figure}[!htbp]
\centering
\includegraphics[width=0.49\textwidth]{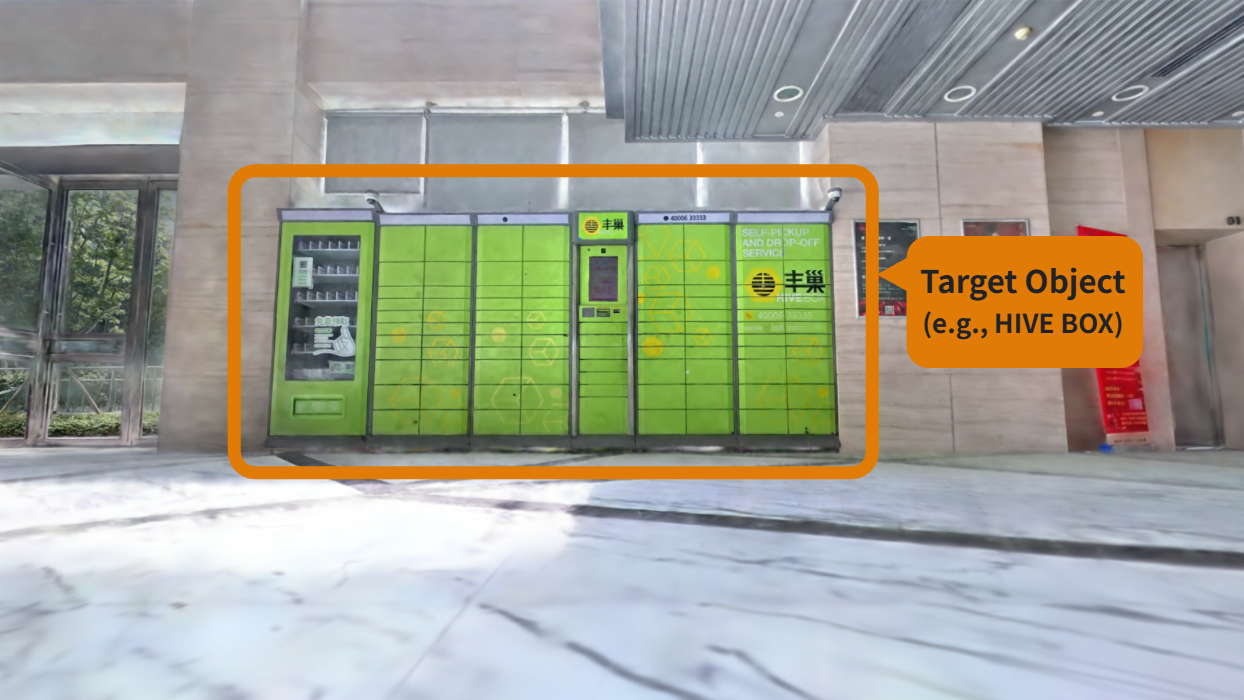}\hfill
\includegraphics[width=0.49\textwidth]{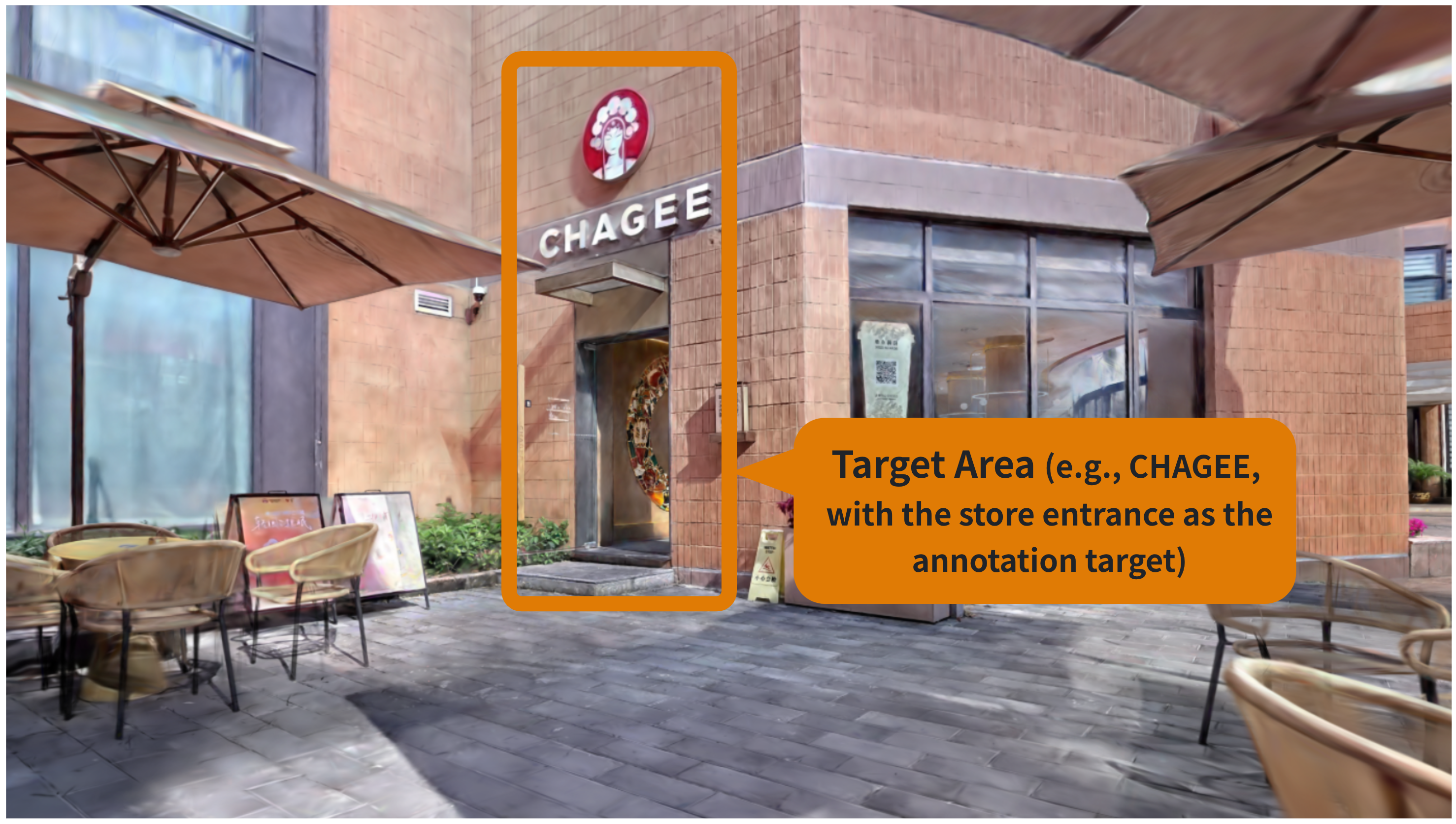}
\caption{Examples of semantic goal annotation. Left: an object target (locker bank). Right: an area target (storefront entrance). Orange outlines mark the annotated targets.}
\label{fig:object-goal-annotation}
\end{figure}

\begin{figure}[!htbp]
\centering
\includegraphics[width=0.72\textwidth,height=0.25\textheight,keepaspectratio]{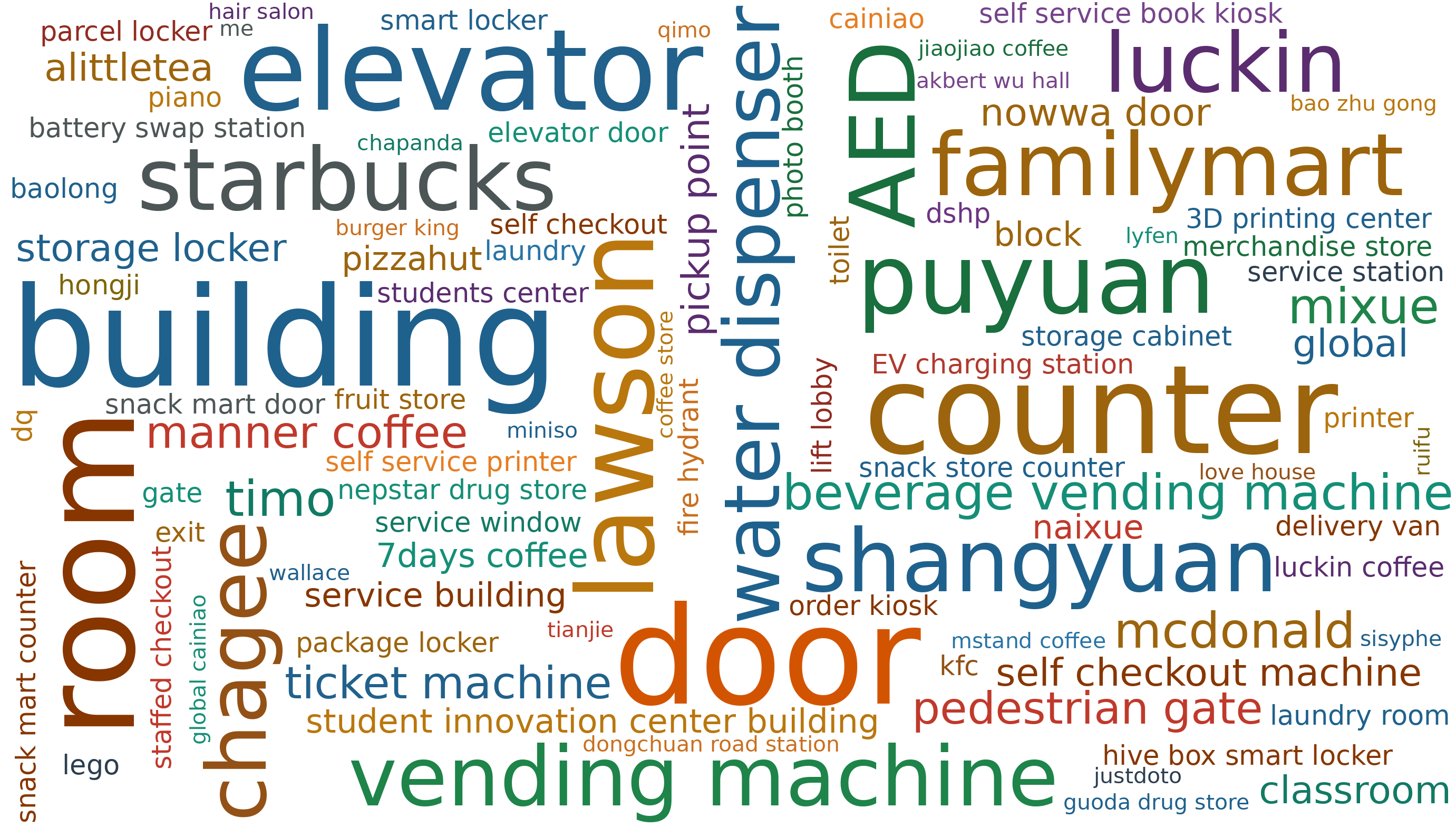}
\caption{Word cloud of annotated OVON goal names. Word size is not interpreted as a measured category or episode frequency.}
\label{fig:object-goal-vocabulary}
\end{figure}

\section{Hierarchical Task Definition}
\label{app:task-definition}
This appendix defines task success, evaluator-gated transitions, and termination. Appendix~\ref{app:metrics} defines metric aggregation.

\subsection{Primitive and Composed Task Structure}
Level~1 evaluates Exit, OutdoorNav, Entrance, and OVON separately. Level~2 evaluates Exit $\to$ OutdoorNav (I2O) or OutdoorNav $\to$ Entrance (O2I). Level~3 evaluates Exit $\to$ OutdoorNav $\to$ Entrance $\to$ OVON. The respective numbers of stages are one, two, and four. The benchmark fixes their order; agent-generated workflow steps do not redefine these stages.

Exit and Entrance contain physical domain crossings. A stage handoff instead changes the active task. Exit and Entrance determine boundary crossing from the agent center: the center must cross into the required region. The indoor/outdoor identity is determined solely by whether the entity is located within the target area.

\subsection{Terminal Events and Success Sets}
\label{app:success-predicates}
Exit, Entrance, and OVON require a valid stop. For these tasks, let $x_i^{\mathrm{stop}}$ be the state at the completion declaration and $z_i$ indicate that such a declaration occurs. The state includes pose, floor, and building context for evaluation. Success requires both the declaration and the task-specific predicate. OutdoorNav instead succeeds upon reaching its goal tolerance on the same navigation plane; no STOP is required.

For tasks requiring STOP, a policy STOP rejected by the method's verifier is not yet a benchmark-stage completion declaration. For a bare policy, its terminal STOP supplies the declaration. For an agent, verifier acceptance supplies it. The evaluator then checks the task predicate independently. Rejection by the verifier can trigger recovery; failure at the evaluator check terminates the episode. A timeout, stationary action, safety brake, or invalid response is not a valid STOP.

Each task has an accepted terminal state set $G_i^q$. This set incorporates the task-specific target and spatial conditions defined below. Boundary tasks also require a valid crossing in the stage trajectory $\tau_i$.

\paragraph{Exit.}
Let $X_i^{\mathrm{I2O}}(\tau_i)$ indicate a valid crossing from the source interior to the outdoor network. Let $G_i^{\mathrm{Exit}}$ contain the allowed outdoor stopping states:
\begin{equation}
S_i^{\mathrm{Exit}}
=z_i X_i^{\mathrm{I2O}}(\tau_i)
\mathbf{1}[x_i^{\mathrm{stop}}\in G_i^{\mathrm{Exit}}].
\end{equation}
The agent center must cross the boundary and lie in an allowed success region. Approaching a doorway is insufficient. The annotation protocol determines which sidewalks, crosswalks, and shared roads are allowed. 
\paragraph{Entrance.}
Let $X_i^{\mathrm{O2I}}(\tau_i)$ indicate entry through a valid boundary into the designated target building. The success set $G_i^{\mathrm{Entrance}}$ contains valid terminal states inside that building:
\begin{equation}
S_i^{\mathrm{Entrance}}
=z_i X_i^{\mathrm{O2I}}(\tau_i)
\mathbf{1}[x_i^{\mathrm{stop}}\in G_i^{\mathrm{Entrance}}].
\end{equation}
This predicate is target-aware. Entry into an arbitrary indoor polygon is not sufficient. Where the task restricts the target floor, the success set must enforce that restriction. Center-point crossing and permitted entry boundaries follow the same annotation convention as Exit.

\paragraph{OutdoorNav.}
For outdoor goal $g_i$, let $d_{xy}$ denote planar Euclidean distance and $P(x,g_i)$ indicate that the agent and goal are on the same navigation plane. We use $\tau_{\mathrm{out}}=3\,\mathrm{m}$ for all reported runs:
\begin{equation}
\begin{aligned}
G_i^{\mathrm{Outdoor}}&=\{x:d_{xy}(x,g_i)\leq\tau_{\mathrm{out}},\ P(x,g_i)=1\},\\
S_i^{\mathrm{Outdoor}}&=\mathbf{1}[\exists t\in[0,T_i]:x_i(t)\in G_i^{\mathrm{Outdoor}}].
\end{aligned}
\end{equation}
Here $x_i(t)$ is the executed state trajectory and $T_i$ is the end of the stage execution. Reaching the accepted set suffices without a STOP. No additional terminal-region, visibility, or reachability check is applied. Goal arrival is distinct from DCR: reaching the goal need not imply that the preceding route stayed in compliant regions.

\paragraph{OVON.}
Let $\mathcal{O}_i$ be the object instances accepted by the language goal, with goal location $g_o$ for each instance $o$. All reported runs use $\tau_{\mathrm{obj}}=3\,\mathrm{m}$ and the same planar-distance and same-plane criteria as OutdoorNav:
\begin{equation}
V_i(o)=\{x:d_{xy}(x,g_o)\leq\tau_{\mathrm{obj}},\ P(x,g_o)=1\}.
\end{equation}
The accepted goal set and success predicate are
\begin{equation}
G_i^{\mathrm{OVON}}=\bigcup_{o\in\mathcal{O}_i}V_i(o),
\qquad
S_i^{\mathrm{OVON}}=z_i\mathbf{1}[x_i^{\mathrm{stop}}\in G_i^{\mathrm{OVON}}].
\end{equation}
Reaching any accepted instance is sufficient. Apart from the same-plane requirement and valid stop, success depends only on distance; no additional target-region, visibility, or reachability check is applied. A target on another floor does not qualify solely because its planar distance is within 3 m.

\subsection{Continuous Execution and Ordered Success}
A handoff occurs only after the evaluator confirms the current stage's success. It retains the actual simulator state, without moving the agent to a sampled start or reference waypoint. After valid Exit completion, the next-stage instruction, outdoor localization, and route become available. A failed Exit declaration releases none of these inputs and terminates the episode.

Let $S_i^{(k)}$ be the evaluator outcome of benchmark stage $k$. A failed stage terminates the chain; unattempted downstream stages are assigned zero. Appendix~\ref{app:metrics} uses the resulting valid prefix for FSR, CSR, conditional success, and failure localization. Workflow steps and verifier rechecks are not additional benchmark stages.

The gate is environment-provided feedback, not evidence that the agent independently judged completion correctly. Evaluator-only region annotations remain hidden from the policy and verifier; the OutdoorNav goal is a task input. For tasks requiring STOP, recovery is allowed within the method's budget before a terminal declaration, but not after evaluator rejection.

\subsection{Failures and Terminal Records}
On timeout or budget exhaustion without valid completion, the unfinished stage fails. The last recorded state and trajectory still support TL, route-progress, and outdoor-compliance calculations. Completed valid prefixes remain credited. No synthetic STOP is added to obtain success.

Simulator, communication, and inference failures are infrastructure errors, distinct from navigation failures. Misalignments caused by infrastructure failures will be re-tested after the batch evaluation finishes.
\section{Episode Generation and Information Conditions}
\label{app:task-generation}
This appendix details episode construction, reference records, input access, and the evaluation-only episode inventory.

\subsection{Episode Specification}
An episode is represented as
\begin{equation}
e=(c,x_0,u,\mathcal{Q},G,B),
\end{equation}
where $c$ is the scene, $x_0$ the initial state, and $u$ the instruction specification. $\mathcal{Q}=(q_1,\ldots,q_K)$ is the predefined ordered stage list. $G$ contains the stage goals and their evaluator conditions. $B$ records the execution limits. $K$ is one for L1, two for L2, and four for L3.

The episode record identifies the scene, initial pose, ordered goals, instructions, and budgets. Reference routes can support validation or scoring; policy access to a route is specified separately. The benchmark supplies $\mathcal{Q}$ rather than requiring autonomous inference of the mission sequence.

\subsection{Connectivity and Candidate Points}
Navigation geometry identifies connected indoor and outdoor regions. Candidate points are grouped as indoor, outdoor, or outdoor boundary-near locations. The task sampler combines these points with object goals and area labels to construct primitive starts and goals. Reference trajectories are generated between candidate starts and accepted goals using scene connectivity. Unreachable candidates are discarded before compatible primitives are composed into longer missions. These references support episode validation and evaluation; designated outdoor routes also serve as task inputs.
The traversable representation is derived using the \texttt{main} branch of PCT planner's point-cloud tomography framework (Appendix~\ref{app:scene-construction}), with a voxel size of $0.1\,\mathrm{m}$. Traversable regions are eroded by $0.4\,\mathrm{m}$ to maintain a minimum obstacle clearance of $0.4\,\mathrm{m}$. Boundary points are sampled in outdoor regions within $15\,\mathrm{m}$ of a building, rather than directly on the indoor--outdoor boundary. Candidate points are initialized by farthest-point sampling (FPS), followed by 10 iterations of Lloyd relaxation to improve spatial coverage.

\paragraph{Point categories.}
For indoor-point sampling, the PCT traversable map is eroded by an additional $0.2\,\mathrm{m}$ after the initial $0.4\,\mathrm{m}$ erosion. Indoor points are sampled from the resulting traversable indoor regions, with at least one point on every floor and the total count set by the scene quota. Outdoor points are sampled from traversable outdoor regions, with 15 points per scene. Boundary points are sampled from traversable outdoor space within $15\,\mathrm{m}$ of an indoor building region, with counts set by the scene quota. FPS followed by Lloyd relaxation is applied to all three categories.

\subsection{Primitive and Composed Episodes}
\paragraph{Exit and Entrance.}
Each sampled indoor point defines an Exit candidate, with the destination specified as an allowed outdoor region. These regions include annotated crosswalks, sidewalks, shared roads, and open areas. Each boundary point defines an Entrance candidate whose goal is entry into the designated indoor building region. Thus, their candidate counts follow the indoor-point and boundary-point counts, respectively. Exit has no additional start--goal distance filter; Entrance uses the boundary-neighborhood restriction above.

\paragraph{OutdoorNav.}
Outdoor points are paired without ordering, so $A\to B$ and $B\to A$ count as one candidate pair. Candidate routes must have a planned path length greater than $20\,\mathrm{m}$. Routes are sampled at evenly spaced positions in distance order to cover short, medium, and long navigation distances within the scene quota.

\paragraph{OVON.}
OVON starts are selected from the sampled indoor points and paired with annotated object goals; starts too close to the goal are excluded. The generation specification restricts each pair to the same building and requires a straight-line separation greater than $5\,\mathrm{m}$. A route is planned for each eligible pair, and final selection uses candidates with valid planned routes. Goal selection must account for the task's accepted terminal set, not only a geometric point.

\paragraph{Composed tasks.}
I2O candidates pair indoor starts with outdoor destinations; O2I candidates pair outdoor starts with indoor destinations; and I2O2I candidates pair sampled indoor starts with annotated object-goal points as final destinations, through the required outdoor segment. Reference routes are planned for candidate pairs and sampled to the scene-specific task quota. The generated reference sequence must satisfy the ordered stage goals, including the final object goal for I2O2I. Runtime handoff poses still come from the agent's actual execution.

\subsection{Reference Paths and Metric Records}
The manifest distinguishes four records: goal annotations, policy-provided outdoor routes, SPL reference lengths, and RC stage-reference curves. Only the designated outdoor route is a task input; evaluator annotations remain hidden.

SPL reference lengths are taken directly from PCT-planned paths, use A* search to find the shortest path, as a shortest-path reference.

Primitive SPL references terminate in the accepted success sets from Appendix~\ref{app:success-predicates}. Exit and Entrance references must include valid boundary crossings. OVON references target a valid stopping state for an accepted instance, not the object's center. A composed reference must visit every required stage goal in order. Concatenated segments need compatible endpoints and valid connectivity; they are not automatically a shortest full-chain route.

RC records contain stage curves, arc-length parameterizations, floor/component labels, and the reference revision. They must specify any connector from the actual handoff pose. In I2O, the provided route starts away from the exit, so its first waypoint receives no automatic progress credit. Reference construction is fixed across methods.

Validation covers success-set reachability, reference continuity, and positive lengths. Timestamped positions, stage events, declarations, and region labels support metric reconstruction. Invalid references and missing traces require explicit reporting.

\subsection{Sampling and Validation}
We randomly generate candidates at three times the target count and uniformly downsample them to a target of 200 episodes per task setting across 20 scenes. Scene quotas are adjusted for scene size, averaging 10 episodes per scene and task. Sampling is stratified across navigation distances, buildings, and building pairs where applicable to reduce concentration on particular routes or locations. A composed mission counts as one episode, not one per constituent stage. The reported evaluation counts are listed in Table~\ref{tab:episode-counts}; the L3 inventory contains 174 episodes, 26 below the target, because some small indoor scenes, such as convenience stores, do not provide enough candidates satisfying the generation rules.

Episode checks cover initial-state validity, connectivity, target identity, and nonempty success sets. Offline connectivity and successful replay through the evaluation motion interface are distinct checks; neither establishes articulated-robot locomotion performance.

The accepted episode list is saved and fixed across methods, which use the same initial states, goals, and instructions. Only valid candidates are retained; rejected-candidate records are not saved. All benchmark episodes are used for evaluation; there is no benchmark training or development split.

\subsection{Instruction and Goal Generation}
Instructions describe primitive goals or complete missions. Predefined next-stage instructions are released after evaluator-confirmed stage completion. This stage handoff is distinct from the agent's within-stage workflow decomposition.

One archived Exit instruction is ``Head out of the building through the exit and stop outdoors. Stay off the grass.'' Its provenance record includes the source file, text hash, selection seed, and instruction index. This example documents an input, not a language-generation model or a complete instruction inventory.

OVON instructions are generated in English by simplifying annotated goal names, followed by a grammar review using Kimi K3. For example, \texttt{shangyuan\_111} is rendered as ``Find room 111,'' and \texttt{starbucks\_counter} as ``Find the Starbucks counter.''

No target reference images are supplied in the reported experiments. OutdoorNav receives a reference route, while OVON instructions use annotated goal names without synonym variants. Instructions are reviewed using Kimi K3.

\subsection{Observations and Information Access}
Visual observations are RGB images, adapted to each navigation policy's native input requirements. No depth observations or target reference images are supplied. OutdoorNav receives accurate simulator-provided position information and the designated reference route. Neither the navigation policies nor the agent can access the full navigation map; maps used for offline episode generation are not online policy inputs.

Controller access to simulator state does not grant the policy or verifier access to success polygons or evaluator building/floor labels. For Exit, Entrance, and OVON, the evaluator checks completion after a terminal declaration; OutdoorNav completion is determined by goal arrival. Its stage gate is the explicit environment feedback described in Appendix~\ref{app:task-definition}; it is not a success signal available for verification or recovery before declaration. 

For tasks beginning with Exit, outdoor localization and the route are released only after evaluator-confirmed Exit success. A failed declaration terminates the episode without releasing them. The I2O route starts away from the building, requiring the agent to connect its exit pose to the route. In O2I, the route reaches the target building's vicinity but does not specify its entrance.

\subsection{Episode Inventory and Coverage Statistics}
\label{app:statistics}
Table~\ref{tab:episode-counts} summarizes the 1,374 evaluation episodes: 800 at L1, 400 at L2, and 174 at L3. These are benchmark episodes, not executions summed over methods or repeated runs. Each result identifies its actual evaluated episode count and any metric-specific eligibility count.
\begin{table}[!htbp]
\centering\small
\setlength{\tabcolsep}{3pt}
\renewcommand{\arraystretch}{1.1}
\caption{Evaluation episode inventory: 200 episodes for each L1 and L2 task and 174 for L3, totaling 1,374. All episodes are used for evaluation.}
\label{tab:episode-counts}
\begin{tabular*}{\textwidth}{@{\extracolsep{\fill}}l*{8}{c}@{}}
\toprule
Level & \multicolumn{4}{c}{L1} & \multicolumn{2}{c}{L2} & L3 & All \\
\cmidrule(lr){2-5}\cmidrule(lr){6-7}\cmidrule(lr){8-8}\cmidrule(lr){9-9}
Task & Exit & OutdoorNav & Entrance & OVON & I2O & O2I & I2O2I & Total \\
\midrule
Episodes & 200 & 200 & 200 & 200 & 200 & 200 & 174 & \textbf{1,374} \\
\bottomrule
\end{tabular*}
\end{table}

Coverage statistics distinguish reference lengths for full missions and individual stages. They also describe scene, building, and floor coverage, together with goal and instruction diversity.
Table~\ref{tab:scene-inventory} lists 252 annotated object goals across 20 scenes and five environment types. These are scene annotations, not distinct object categories or episode-level target frequencies.

\section{Hierarchical Metrics and Failure Localization}
\label{app:metrics}
This appendix defines the metrics in Secs.~\ref{sec:level1}--\ref{sec:level3}. Success predicates and termination follow Appendix~\ref{app:success-predicates}. Rate and progress metrics lie in $[0,1]$ and are displayed as percentages; TL is measured in meters.

\subsection{Notation and Valid Prefixes}
For one task and method configuration, let $N=N_0$ be the number of admitted episodes. Let $K$ be the number of required stages: $K=2$ for I2O/O2I and $K=4$ for I2O2I. The binary outcome $S_i^{(k)}$ denotes valid completion of stage $k$; an unattempted stage is assigned zero. Define
\begin{equation}
A_i^{(0)}=1,\qquad
A_i^{(k)}=\prod_{j=1}^{k}S_i^{(j)},\qquad
N_k=\sum_{i=1}^{N}A_i^{(k)}.
\label{eq:valid-prefix}
\end{equation}
$A_i^{(k)}=1$ means that the first $k$ stages were completed validly in order. Thus $N_0\geq N_1\geq\cdots\geq N_K$. The valid completed-prefix length is
\begin{equation}
c_i=\sum_{k=1}^{K}A_i^{(k)}.
\end{equation}
The evaluator terminates a chain at its first failed stage. Unattempted stages receive no credit, and an agent's internal stage index does not replace the evaluator's prefix count.

\subsection{L1 Success, Trajectory Length, and Path Efficiency}
Let $S_i$ be the task-success indicator and $P_i$ the total executed path length. Let $L_i$ be the reference length taken directly from the PCT-planned path. Then
\begin{equation}
\mathrm{SR}=\frac{1}{N}\sum_i S_i,\qquad
\mathrm{TL}=\frac{1}{N}\sum_i P_i,\qquad
\mathrm{SPL}=\frac{1}{N}\sum_i S_i\frac{L_i}{\max(L_i,P_i)}.
\label{eq:primitive-metrics}
\end{equation}
Exit, Entrance, and OVON report these three metrics. OutdoorNav reports SR and SPL together with DCR; its TL may be reported as a supplementary statistic. TL averages all admitted episodes. It is not ranked as lower-is-better because an early failure can have little travel.

For Exit and Entrance, the reference must reach an accepted terminal region through a valid boundary crossing. For OVON, it must reach the closest valid stopping state associated with any accepted object instance. For OutdoorNav, it must reach the declared goal-tolerance set. 

Executed and reference lengths use the same geometry and distance convention. $P_i$ includes verification and recovery movement. The PCT reference is used without additional smoothing; the reported SPL is normalized by a planner reference . With common episodes and valid denominators, $\mathrm{SPL}\leq\mathrm{SR}$.

\subsection{Full Success and Prefix Completion}
Full success rate (FSR) and completion success rate (CSR) are
\begin{equation}
\mathrm{FSR}=\frac{N_K}{N}
=\frac{1}{N}\sum_i A_i^{(K)},\qquad
\mathrm{CSR}=\frac{1}{NK}\sum_i\sum_{k=1}^{K}A_i^{(k)}
=\frac{1}{N}\sum_i\frac{c_i}{K}.
\label{eq:composition-success}
\end{equation}
FSR requires both L2 stages or all four L3 stages. CSR credits only the valid completed prefix. Per-episode CSR is $0,1/2,1$ for L2 and $0,1/4,1/2,3/4,1$ for L3. Mean completed stages equals $K\,\mathrm{CSR}$.

For composed tasks, replace the primitive indicator in SPL with $F_i=A_i^{(K)}$:
\begin{equation}
\mathrm{SPL}_{\mathrm{chain}}
=\frac{1}{N}\sum_i F_i\frac{L_i}{\max(L_i,P_i)}.
\label{eq:chain-spl}
\end{equation}
$P_i$ covers the executed episode, including retries and handoffs. $L_i$ is the PCT-planned reference length for the ordered mission, without independent shortest-path validation. A direct route to the final mission goal may bypass a required stage and is not a valid substitute. Stage references must join consistently; independently minimized segments need not form a feasible or shortest full chain. Failed episodes contribute zero. The mean is over all $N$ episodes, not only successful ones, so $\mathrm{SPL}_{\mathrm{chain}}\leq\mathrm{FSR}$.

\subsection{Reference-Route Completion}
\label{app:route-completion}
Route completion (RC) is reported for the chained tasks (L2 and L3), for which
an episode comprises a sequence of benchmark stages. RC quantifies the extent
to which the chain advanced along its reference route, expressed as an arc
length along the reference polyline, and is defined for successful and failed
episodes alike.

\paragraph{Reference polyline.}
For episode $i$, let $R_i=(p_{i,0},\dots,p_{i,M})$ denote the reference route
of the task represented as a polyline, with horizontal length
$L_i=\sum_{j}\|p_{i,j+1}-p_{i,j}\|>0$. $R_i$ is the complete start-to-goal
route of the task, i.e.\ the top-level \texttt{trajectory} of the task record;
for L3 it corresponds to the full route through the building. A single
polyline thus represents an episode.

\paragraph{Projection of the terminal position.}
Let $q_i$ denote the terminal recorded position of the chain, defined as the
last point of its last non-empty stage. The projection of $q_i$ is computed
over all segments of $R_i$. For the segment $[p_{i,j},p_{i,j+1}]$, the
truncated projection parameter is
\begin{equation}
t_j=\mathrm{clip}_{[0,1]}\!\left(
\frac{(q_i-p_{i,j})\cdot(p_{i,j+1}-p_{i,j})}
     {\|p_{i,j+1}-p_{i,j}\|^2}\right),
\end{equation}
with foot point $p_{i,j}+t_j\,(p_{i,j+1}-p_{i,j})$. The foot point minimising
the distance to $q_i$ is retained (ties resolved in favour of the smaller
segment index), and
\begin{equation}
s_i=\sum_{l<j}\|p_{i,l+1}-p_{i,l}\|+t_j\,\|p_{i,j+1}-p_{i,j}\|
\label{eq:route-projection}
\end{equation}
denotes the arc length from $p_{i,0}$ to that foot point. Accordingly,
\begin{equation}
\mathrm{RC}_i=\mathrm{clip}_{[0,1]}\!\left(s_i/L_i\right),
\qquad
\mathrm{RC}=\frac{1}{|\mathcal{I}|}\sum_{i\in\mathcal{I}}\mathrm{RC}_i,
\label{eq:route-completion}
\end{equation}
where $\mathcal{I}$ denotes the set of episodes with non-degenerate reference
($L_i>0$). Equivalently, $\mathrm{RC}_i$ is the fraction of the reference
lying between its origin and the point of the reference closest to $q_i$.

\paragraph{Properties.}
(i) \emph{RC is a function of the terminal position alone.} A chain is
evaluated by its point of termination.
(ii) \emph{RC is a planar geometric quantity.} It is defined on the horizontal
($xy$) coordinates of the terminal position and of the reference polyline.
(iii) \emph{Progress is weighted by route length.} Since the reference is a
single polyline, the RC increment per unit distance travelled is uniform along
the route, and the contribution of a stage is proportional to its length.
(iv) \emph{RC is defined for every episode} and is reported alongside the
success verdict rather than being subsumed by it. On the chained tasks,
successful episodes attain a mean $\mathrm{RC}\approx 0.90$--$0.98$, of which
$7$--$62\%$ equal $1$ exactly, depending on task and policy; failed chains
attain partial values, with means of $0.18$--$0.82$ for the L2 and L3
baselines. For L3 the reference extends into the building, so RC reports the
position attained along the complete route.
(v) \emph{RC is truncated to $[0,1]$.} Positions beyond the route endpoints,
or off the route, are mapped into that interval.

\subsection{Outdoor-Only Distance Compliance}
\label{app:outdoor-dcr}
Distance compliance rate (DCR) measures static preferred-region compliance. 

Let $\mathcal{T}_i^{\mathrm{out}}$ contain all physically outdoor trajectory segments across the episode, including outdoor portions of Exit, Entrance, and any other stage. Membership depends on location, not the active stage. Each segment has length $\ell_t$ and representative position $m_t$. The binary label $b(m_t)$ is one in a compliant region and zero in a noncompliant one. Then
\begin{equation}
D_i^{\mathrm{out}}=\sum_{t\in\mathcal{T}_i^{\mathrm{out}}}\ell_t,\qquad
\mathrm{DCR}_i=
\frac{\sum_{t\in\mathcal{T}_i^{\mathrm{out}}}\ell_t b(m_t)}
     {D_i^{\mathrm{out}}},\quad D_i^{\mathrm{out}}>0.
\label{eq:outdoor-dcr}
\end{equation}
Indoor movement never enters either sum. Split segments at indoor/outdoor boundaries; use a fixed subdivision rule at compliance boundaries. Compliant sidewalks, crosswalks, and pedestrian-permitted shared roads follow the reviewed annotation. Final access labels are resolved by human annotators; no uncertain category remains. For overlapping regions, if both compliant and non-compliant region annotations exist, the non-compliant region division is preferred so as to guarantee compliance.

Failed episodes remain eligible when they contain nonzero outdoor travel. If $D_i^{\mathrm{out}}=0$, DCR is undefined, not zero or one. With
\begin{equation}
\mathcal{E}=\{i:D_i^{\mathrm{out}}>0\},\qquad
N_{\mathrm{eligible}}=|\mathcal{E}|,
\end{equation}
the reported episode-average metric is
\begin{equation}
\mathrm{DCR}=
\frac{1}{N_{\mathrm{eligible}}}\sum_{i\in\mathcal{E}}\mathrm{DCR}_i,
\quad N_{\mathrm{eligible}}>0.
\end{equation}
Report $N_{\mathrm{eligible}}$ and the OutdoorNav-stage reach count alongside the aggregate. Physical outdoor travel and activation of the OutdoorNav stage are different events. A separately reported distance-pooled DCR weights long trajectories more heavily and must not use the same unlabeled metric name.
\subsection{Stage-Conditional Success and Failure Localization}
The cumulative prefix-completion rate is $N_k/N$. Stage-conditional success is
\begin{equation}
C_k=\frac{N_k}{N_{k-1}},\qquad N_{k-1}>0.
\label{eq:stage-conditional}
\end{equation}
For I2O, $(C_1,C_2)$ denotes Exit success and OutdoorNav success conditioned on valid Exit. For O2I, it denotes OutdoorNav success and Entrance success conditioned on valid OutdoorNav. For L3, the four rates follow Exit, OutdoorNav, Entrance, and OVON. Each denominator includes all preceding stages, not only the immediately preceding local outcome.

A zero denominator makes a rate undefined, not zero. Report $N_0,\ldots,N_K$ alongside the rates. If all conditional denominators are positive, the ratios telescope to $\mathrm{FSR}=N_K/N=\prod_{k=1}^{K}C_k$. This counting identity assumes no independence. If a prefix count is zero, FSR is zero but later conditional rates remain undefined.

For a failed episode, the first failed stage is $c_i+1$. Full successes have no failed stage. Among admitted episodes with complete outcome records, the count first failing at stage $k$ is $N_{k-1}-N_k$. Report whether proportions use all admitted episodes or only failed episodes. Do not count unattempted downstream stages as additional first failures. Infrastructure errors require a separate label and a declared inclusion/rerun rule.

\subsection{Primitive-Product Reference and Matched Comparison}
Let $r_q=\mathrm{SR}^{L1}_q$ be the standalone primitive success fraction for the matching policy and information condition. Define
\begin{equation}
R_{\mathrm{I2O}}=r_{\mathrm{Exit}}r_{\mathrm{OutdoorNav}},\qquad
R_{\mathrm{O2I}}=r_{\mathrm{OutdoorNav}}r_{\mathrm{Entrance}},
\end{equation}
\begin{equation}
R_{\mathrm{I2O2I}}=
r_{\mathrm{Exit}}r_{\mathrm{OutdoorNav}}r_{\mathrm{Entrance}}r_{\mathrm{OVON}}.
\end{equation}
The products use fractions, not percentages. They are independence-based references, not predictions or bounds on composed success. L1 and composed tasks differ in starting states, difficulty, history, and budget use; a difference from the product alone does not identify a composition penalty.

Matched comparisons align episode IDs, goals, executor assignments, inputs, stopping rules, and budgets. Composed runs preserve actual handoff states; prescribed-entry and handoff-replay diagnostics are separate conditions. Conditional stage rates describe selected subsets, so they are not directly interchangeable with standalone L1 rates. A matched action cap also does not imply matched token use or latency.

Equal-budget continuation isolates additional execution from semantic verification. Its continuation and stopping rules must be fixed without querying hidden success annotations or tuning on the evaluation episodes.

\subsection{Failure Labels and Qualitative Cases}
Failure labels distinguish primitive execution, stopping, handoffs, and recovery. For Exit, Entrance, and OVON, a no-stop label requires evidence of a valid goal state without a completion declaration; this label does not apply to OutdoorNav. Next-stage failure alone does not establish a bad handoff. Infrastructure-error accounting follows Appendix~\ref{app:task-definition}.
The audit covers 31 purposefully selected cases, including failures and successful comparisons. Twenty-seven have visual evidence (115 selected images); four have only logs and trajectories. This nonrandom sample illustrates behaviors, not failure-type frequencies.

Figure~\ref{fig:primitive-failures} uses C04 (Uni-NaVid, scene 01, task 0), C06 (agent, scene 12, task 8), C12 (OmniNav, scene 01, task 0), and C16 (agent, scene 12, task 0). The manifest links images to results, original frames, steps, and source hashes. Frame intervals are unequal. 

\section{Simulation and Embodiment Support}
\label{app:simulation}
This appendix distinguishes available simulation capabilities from the execution mode used in evaluation.

\subsection{Simulation Backends and Scene Assets}
The benchmark supports Isaac Sim and Habitat, with Gaussian rendering in Habitat provided through Habitat-GS~\cite{habitatgs}. Scene alignment follows Appendix~\ref{app:scene-construction}. The reported experiments use Isaac Sim.
\subsection{Kinematic Execution and Robot Support}
To reduce the influence of unreliable locomotion control, the reported evaluation uses a kinematic agent with collision constraints in Isaac Sim. It carries sensors and is prevented from passing through scene geometry. Traversable stairs can be ascended or descended without an articulated locomotion policy. This setting evaluates navigation decisions under the declared motion and collision interface; it does not establish articulated-robot balance, contact dynamics, or locomotion performance.

Isaac Sim implementations include M20, A2W, and Go2W robot models with configurable sensors. These models have been tested in a simulation environment and exhibit the corresponding motion and contact-dynamics performance. Figure~\ref{fig:robot-stage-examples} illustrates a supported robot embodiment across the four navigation stages.

\begin{figure}[!htbp]
\centering
\begin{minipage}[t]{0.49\linewidth}
\centering
\includegraphics[width=\linewidth]{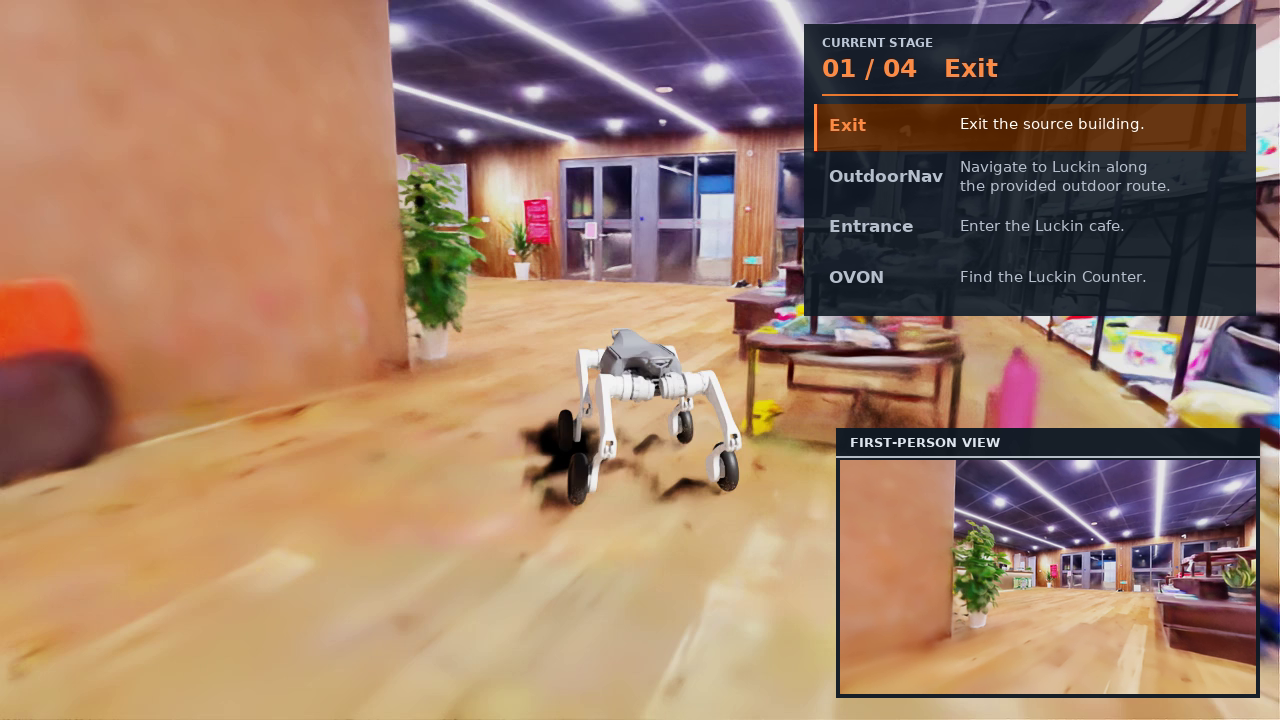}\\[2pt]
{\small (a) Exit}
\end{minipage}\hfill
\begin{minipage}[t]{0.49\linewidth}
\centering
\includegraphics[width=\linewidth]{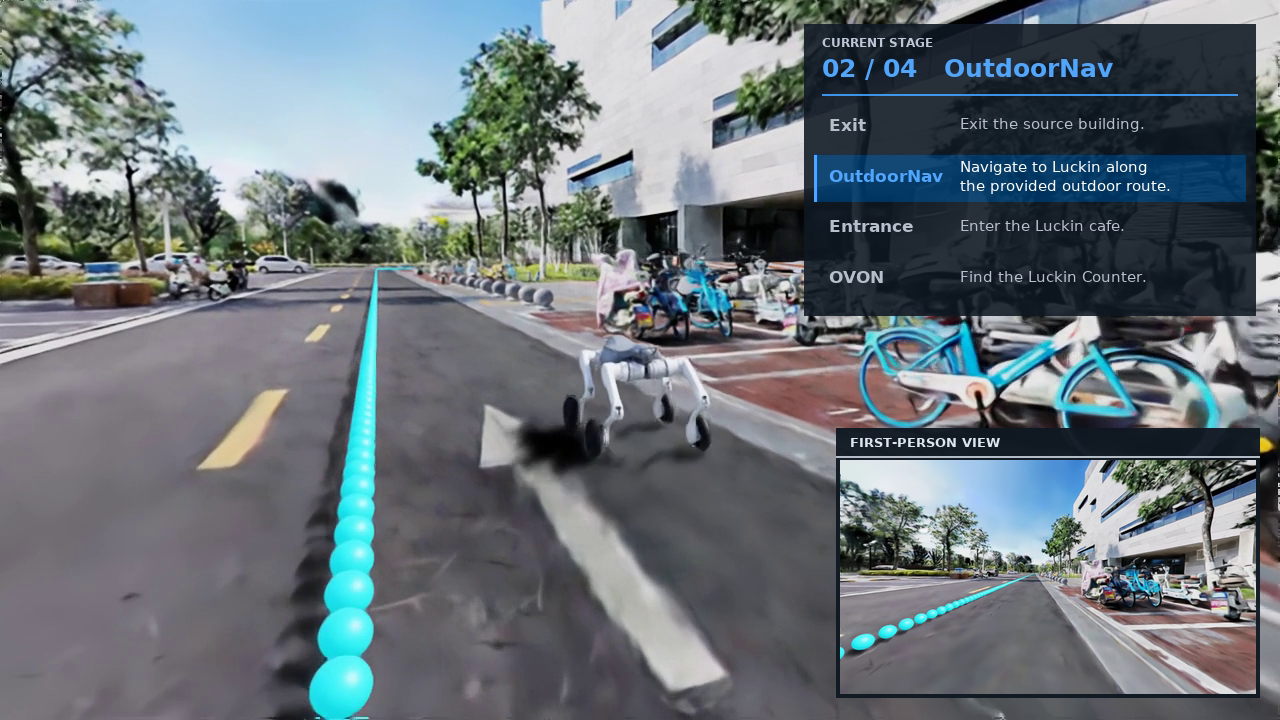}\\[2pt]
{\small (b) OutdoorNav}
\end{minipage}
\par\vspace{6pt}
\begin{minipage}[t]{0.49\linewidth}
\centering
\includegraphics[width=\linewidth]{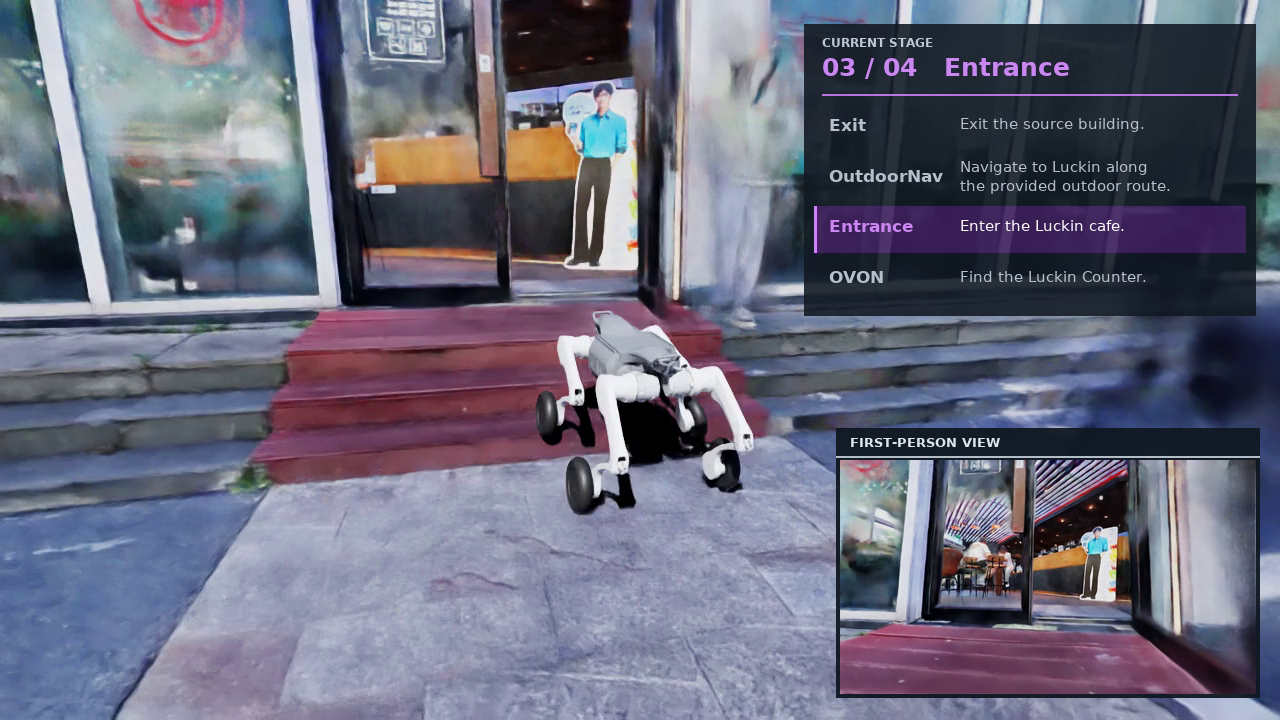}\\[2pt]
{\small (c) Entrance}
\end{minipage}\hfill
\begin{minipage}[t]{0.49\linewidth}
\centering
\includegraphics[width=\linewidth]{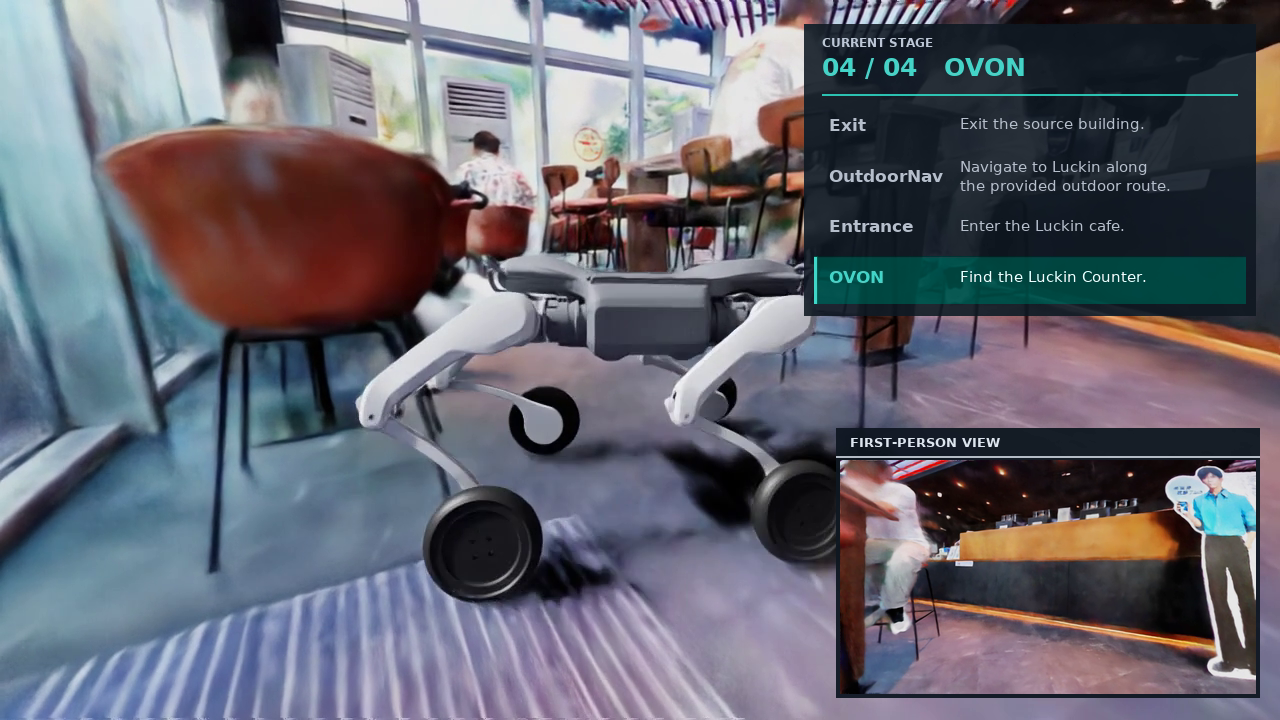}\\[2pt]
{\small (d) OVON}
\end{minipage}
\caption{Robot embodiment across the four I2O2I navigation stages.}
\label{fig:robot-stage-examples}
\end{figure}

\subsection{Sensors, Motion Interface, and Run Configuration}
\label{app:interface}

All reported experiments are conducted in Isaac Sim 6.0.1 using a collision-constrained kinematic agent. Camera settings, image preprocessing, input views, and observation-history handling follow the requirements of each navigation policy's official open-source implementation. Policy outputs are executed through policy-specific adapters following the corresponding action conventions. For comparisons between the fixed policy chain and MODULAR-Agent, the underlying navigation policies and their input configurations are kept unchanged. Task-specific information access follows Appendix~\ref{app:task-generation}.

\section{Transition-Aware Reference Agent}
\label{app:agent-details}
Success follows Appendix~\ref{app:task-definition}; matched comparisons follow Appendix~\ref{app:metrics}.

\subsection{Composable Workflows and Stage Management}
The agent composes within-stage steps, such as finding and descending stairs. Steps retain the parent goal and simulator state; their completion leaves the benchmark stage unchanged. A single loop dispatches the skills: location assessment, subtask planning, and completion verification. The planning skill also serves replanning when a replan context is supplied. The subtask queue and its completion predicates come from the task-state module, whose stage order fixes the mission sequence for each composition level.

The verifier checks completion using allowed observations. Rejection can trigger bounded recovery. An accepted stage proposal reaches the evaluator: success advances the mission; failure terminates it without further recovery (Appendix~\ref{app:task-definition}).

\subsection{Base Navigation Policies and Adapters}
In L1, (O) uses OmniNav, with the OmniNav-fast configuration for OutdoorNav; (S) uses SocialNav for OutdoorNav. In composed (O + S) runs, OmniNav executes Exit, Entrance, and OVON, and SocialNav executes OutdoorNav. The fixed chain uses the same assignments. These evaluated configurations form a subset of the policy library. Table~\ref{tab:policy-registry} lists the registry entries, their accepted stage names, and the role each entry plays in the reported runs.

\begin{table}[htbp]
\centering
\small
\begin{tabular}{@{}p{0.15\textwidth}p{0.25\textwidth}p{0.17\textwidth}p{0.28\textwidth}@{}}
\toprule
Policy & Registry entry and aliases & Stages & Role \\
\midrule
OmniNav & omninav (omninav\_3view) & exit, entrance, ovon & default executor for Exit, Entrance, OVON \\
OmniNav-fast & omninav\_fast (omninav-fast) & exit, entrance, outdoornav & outdoor executor in the (O) configuration \\
SocialNav & socialnav (social-nav) & outdoornav & outdoor executor in the (S) configuration \\
Uni-NaVid & uninavid (uni-navid) & exit, entrance, ovon & baseline policy \\
LightNav-0 & lightnav (lightnav0) & exit, entrance, ovon & baseline policy \\
\bottomrule
\end{tabular}
\caption{Policy registry entries used by the evaluation. Registry status gates an entry at build time.}
\label{tab:policy-registry}
\end{table}

The stage-to-executor assignment is a default of the task profile: Exit, Entrance, and OVON take OmniNav, and OutdoorNav takes SocialNav. A phase-policy override substitutes any registered policy for a named stage, and the composed evaluator forwards the outdoor override to the agent. The name \emph{OmniNav-fast} denotes the fast branch of the OmniNav service rather than a separate policy.

Each registry entry is a client to a remote model server: the adapter holds a server URL, a timeout, and stopping parameters, so the evaluated policies run frozen weights that remain on the serving host. The repository carries no navigation checkpoint.

Adapters preserve action semantics, frames, units, horizons, and timing. Absolute waypoints are not incremental displacements.

\subsection{Execution Skills and Completion Verification}
Assessment and decomposition guide execution; verification and replanning respect the task's input and budget limits.

Boundary verification distinguishes approach, crossing, and stopping; object verification follows the language goal. Neither uses hidden evaluator annotations. Table~\ref{tab:verifier-config} records the verifier configuration for the reported runs.

\begin{table}[htbp]
\centering
\small
\begin{tabular}{@{}p{0.24\textwidth}p{0.68\textwidth}@{}}
\toprule
Item & Setting \\
\midrule
Model & kimi-k2.6 \\
Endpoint & /chat/completions under the configured base URL https://api.kimi.com/coding/v1 \\
Decoding & thinking disabled; max\_tokens $256$ for location judgment, $1024$ for planning and replanning, $512$ for verification; request timeout $60$\,s \\
Images & Up to four RGB frames in fixed order: the step-start view, then the forward, left, and right views captured after the stop; JPEG with long edge $640$ \\
Messages & One system prompt and one user message per call, with prior events entering as rendered text \\
Parser fields & step\_completed, step\_confidence, step\_reason, parent\_completed, parent\_confidence, parent\_reason, evidence, retry\_guidance \\
Acceptance & Verdict confidence at or above $0.7$ \\
\bottomrule
\end{tabular}
\caption{Verifier configuration for the reported runs.}
\label{tab:verifier-config}
\end{table}

The parser accepts JSON booleans for the two completion fields and finite confidences in $[0,1]$, and it reports a verdict only after both checks pass. OutdoorNav under SocialNav or the OmniNav-fast configuration uses the deterministic geometric check outdoor\_geometric\_v1, whose confidence takes the values $1$ and $0$.

We distinguish ACCEPT, REJECT, and UNCERTAIN as logical outcomes, not literal parser tokens. A malformed response consumes the attempt and enters the record as a verification-schema failure. A provider-side failure enters the episode record as a system error, which the scorer excludes from its denominator and reports alongside the scored counts.

\subsection{Memory, Recovery, and Budgets}
Each parent workflow allows up to three replanning cycles in addition to the initial plan, yielding at most four execution attempts. The replanning budget is shared across subtasks within the workflow and resets at each benchmark-stage boundary. Failed verification triggers replanning while budget remains; otherwise, the agent reports workflow failure. Recovery takes place before the evaluator's terminal check and remains subject to the cumulative budgets in Table~\ref{tab:agent-budgets}.

\begin{table}[htbp]
\centering
\small
\begin{tabular}{@{}lll@{}}
\toprule
Budget & Value & Scope \\
\midrule
Executed actions & $800$ & episode \\
Steps per attempt & $200$ & attempt \\
Wall clock & $900$\,s & episode \\
Model requests & $50$ & episode \\
Verification cadence & every $50$ steps & stage \\
Periodic verifications & $20$ & episode \\
Progress checkpoints & $3$ & stage \\
Verification threshold & $0.7$ & verdict \\
\bottomrule
\end{tabular}
\caption{Execution and recovery budgets of the reference agent. Model requests count actual HTTP calls.}
\label{tab:agent-budgets}
\end{table}

A progress checkpoint fires when a window of $16$ consecutive actions produces a net translation below $0.5$\,m together with an absolute rotation of at least $\pi/2$ radians. Periodic verification follows the stated cadence for every stage other than OutdoorNav, whose completion the geometric check decides. Both paths increment the same counter and reach the same advance gate.

Agent memory and policy-local history are separate. Two memory objects persist: an episode memory per stage and a mission memory per episode. At a stage handoff the mission memory retains the ordered task contract, the outcomes sourced from earlier stages, and the failure hypotheses, while it invalidates direction facts together with the recent observation and execution evidence. The policy client resets before each step, so policy-local history covers a single step.

Recovery revises instructions or gathers evidence without resetting the goal or state.

\subsection{Verifier Audit and Resource Use}
\label{app:additional-experiments}
Verifier labels use the independent success predicates in Appendix~\ref{app:success-predicates}. Workflow-step and benchmark-stage proposals are audited separately. False stage acceptance leads to evaluator rejection, not a handoff. For $n_{\mathrm{invalid}}$ invalid and $n_{\mathrm{valid}}$ valid proposals, define
\begin{equation}
\mathrm{FAR}=\frac{n_{\mathrm{accepted,invalid}}}{n_{\mathrm{invalid}}},
\qquad
\mathrm{FRR}=\frac{n_{\mathrm{rejected,valid}}}{n_{\mathrm{valid}}}.
\end{equation}
UNCERTAIN is separate from REJECT; zero denominators remain undefined. Proposal-level error rates are not episode-level failure rates. The agent records every verification verdict together with its source, which the policy stop signal, the geometric check, and the periodic check identify, and it records every attempt outcome; these records supply the proposal-level counts that FAR and FRR require.

\clearpage
\endgroup

\end{document}